\documentclass{article}

\usepackage{amsmath,amsfonts,bm}

\def\eqref#1{equation~\ref{#1}}

\def\1{\bm{1}}

\def\vb{{\bm{b}}}

\def\vh{{\bm{h}}}

\def\vl{{\bm{l}}}

\def\vx{{\bm{x}}}

\def\mW{{\bm{W}}}

\DeclareMathAlphabet{\mathsfit}{\encodingdefault}{\sfdefault}{m}{sl}
\SetMathAlphabet{\mathsfit}{bold}{\encodingdefault}{\sfdefault}{bx}{n}
\newcommand{\tens}[1]{\bm{\mathsfit{#1}}}

\def\tL{{\tens{L}}}

\def\tW{{\tens{W}}}

\newcommand{\R}{\mathbb{R}}

\usepackage{amsmath}
\usepackage{hyperref}
\usepackage{url}
\usepackage{graphicx}
\usepackage{subfig}
\usepackage{float}
\usepackage{wrapfig}
\usepackage{chngcntr}
\usepackage{xr}
\usepackage{microtype}
\usepackage{booktabs} 

\usepackage[accepted]{icml2020}

\begin{document}

\twocolumn[
\icmltitle{Walking the Tightrope: An Investigation of the Convolutional Autoencoder Bottleneck}

\begin{icmlauthorlist}
	\icmlauthor{Ilja Manakov}{dbs,eye}
	\icmlauthor{Markus Rohm}{dbs,eye}
	\icmlauthor{Volker Tresp}{dbs,siemens}
\end{icmlauthorlist}

\icmlaffiliation{dbs}{Chair for Database Systems and Data Mining, Ludwig Maximilian University, Munich, Germany}
\icmlaffiliation{eye}{Department of Ophthalmology, Ludwig Maximilian University, Munich, Germany}
\icmlaffiliation{siemens}{Siemens AG, Munich, Germany}

\icmlcorrespondingauthor{Ilja Manakov}{ilja.manakov@gmx.de}

\icmlkeywords{Autoencoders, Unsupervised Learning, Computer Vision}

\vskip 0.3in
]

\printAffiliationsAndNotice{} 

\newcommand{\fix}{\marginpar{FIX}}
\newcommand{\new}{\marginpar{NEW}}
\newcommand*\rfrac[2]{{}^{#1}\!/_{#2}}

\begin{abstract}
In this paper, we present an in-depth investigation of the convolutional autoencoder (CAE) bottleneck.
Autoencoders (AE), and especially their convolutional variants, play a vital role in the current deep learning toolbox.
Researchers and practitioners employ CAEs for a variety of tasks, ranging from outlier detection and compression to transfer and representation learning.
Despite their widespread adoption, we have limited insight into how the bottleneck shape impacts the emergent properties of the CAE.
We demonstrate that increased height and width of the bottleneck drastically improves generalization in terms of reconstruction error while also speeding up training.
The number of channels in the bottleneck, on the other hand, is of secondary importance.
Furthermore, we show empirically that, contrary to popular belief, CAEs do not learn to copy their input, even when all layers have the same number of neurons as there are pixels in the input.
Copying does not occur, even when training the CAE for 1,000 epochs on a tiny ($\approx$ 600 images) dataset.
Besides raising important questions for further research, our findings are directly applicable to two of the most common use-cases for CAEs:
In image compression, it is advantageous to increase the feature map size in the bottleneck as this improves reconstruction quality greatly.
For reconstruction-based outlier detection, we recommend decreasing the feature map size so that out-of-distribution samples will yield a higher reconstruction error.
\end{abstract}

\section{Introduction}
Autoencoders (AE) are an integral part of the neural network toolkit.
They are a class of neural networks that consist of an encoder and decoder part and are trained by reconstructing datapoints after encoding them.
Due to their conceptual simplicity, autoencoders often appear in teaching materials as introductory models to the field of unsupervised deep learning.
Nevertheless, autoencoders have enabled major contributions in the application and research of the field.
The main areas of application include outlier detection \citep{xia15, chen17, zhou17, baur19}, data compression \citep{yildirim18, cheng18, dumas18}, and image enhancement \citep{mao16, lore17}.
In the early days of deep learning, autoencoders were an indispensable catalyst in the training of deep models.
Training large (by the standards of the time) models was challenging, due to the lack of big datasets and computational resources.
One way around this problem was to pre-train some or all layers of the network greedily by treating them as autoencoders with one hidden layer \citep{bengio07}.
Subsequently, \citet{erhan09} demonstrated that autoencoder pre-training also benefits generalization.
Currently, researchers in the field of representation learning frequently rely on autoencoders for learning nuanced and high-level representations of data \citep{kingma13, tretschk19, shu18, makhzani15, berthelot18}.

However, despite its widespread use, we propose that the (deep) autoencoder model is not well understood.
Many papers have aimed to deepen our understanding of the autoencoder through theoretical analysis \citep{nguyen18, arora13, baldi12, alain12}.
While such analyses provide valuable theoretical insight, there is a significant discrepancy between the theoretical frameworks and actual behavior of autoencoders in practice, mainly due to the assumptions made (e.g., weight tying, infinite depth) or the simplicity of the models under study.
Others have approached this issue from a more experimental angle \citep{arpit15, bengio13, le13, vincent08, berthelot19}.
Such investigations are part of an ongoing effort to understand the behavior of autoencoders in a variety of settings.

The focus of most such investigations so far has been the traditional autoencoder setting with fully connected layers.
When working with image data, however, the default choice is to use convolutions, as they provide a prior that is well suited to this type of data \citep{ulyanov18}.
For this reason, \citet{masci11} introduced the convolutional autoencoder (CAE) by replacing the fully connected layers in the classical AE with convolutions.
In an autoencoder, the layer with the least amount of neurons is referred to as a bottleneck.
In the regular AE, this bottleneck is simply a vector (rank-1 tensor).
In CAEs, however, the bottleneck assumes the shape of a multichannel image (rank-3 tensor, height $\times$ width $\times$ channels) instead.
This bottleneck shape prompts the question: What is the relative importance of the number of channels versus the height and width (hereafter referred to as size) in determining the tightness of the CAE bottleneck?
Intuitively, we might expect that only the total number of neurons should matter since convolutions with one-hot filters can distribute values across channels.
Generally, the study of CAE properties appears to be underrepresented in literature, despite their widespread adoption.

In this paper, we share new insights into the properties of convolutional autoencoders, which we gained through extensive experimentation.
We address the following questions:
\begin{itemize}
	\item{How do the number of channels and the feature map size in the bottleneck layer impact}
	\begin{itemize}
		\item{reconstruction quality?}
		\item{generalization ability?}
		\item{the structure of the latent code?}
		\item{knowledge transfer to downstream tasks?}
	\end{itemize}
	\item{How and when do CAEs overfit?}
	\item{How does the complexity of the data distribution affect all of the above?}
	\item{Are CAEs capable of learning a ``copy function'' if the CAE is complete (i. e., when the number of pixels in input equals the number of neurons in bottleneck)? This ``copying CAE'' hypothesis is a commonly held belief that was carried over from regular AEs (see Sections 4 and 5 in \citet{masci11}.}
\end{itemize}

We begin the following section by formally introducing convolutional autoencoders and explaining the convolutional autoencoder model we used in our experiments.
Additionally, we introduce our three datasets and the motivation for choosing them.
In Section \ref{experiments}, we outline the experiments and their respective aims.
Afterward, we present and discuss our findings in Section \ref{results}.
All of our code, as well as the trained models and datasets, will be published \href{https://github.com/IljaManakov/WalkingTheTightrope}{on github}.
This repository will also include an interactive Jupyter Notebook for investigating the trained models.
We invite interested readers to take a look and experiment with our models.
\section{Materials and Methods} \label{methods}
\subsection{Autoencoders and Convolutional Autoencoders}
The regular autoencoder, as introduced by \citet{rumelhart85}, is a neural network that learns a mapping from data points in the input space $\vx \in \R^d$ to a code vector in latent space $\vh \in \R^m$ and back.
Typically, unless we introduce some other constraint, $m$ is set to be smaller than $d$ to force the autoencoder to learn higher-level abstractions by having to compress the data.
In this context, the encoder is the mapping $f(\vx): \R^d \to \R^m$ and the decoder is the mapping $g(\vh): \R^m \to \R^d$.
The layers in both the encoder and decoder are fully connected:
\begin{equation}
	\vl^{i+1} = \sigma(\mW^i\vl^i + \vb^i).
\end{equation}
Here, $\vl^i$ is the activation vector in the i-th layer, $\mW^i$ and $\vb^i$ are the trainable weights and $\sigma$ is an element-wise non-linear activation function.
If necessary, we can tie weights in the encoder to the ones in the decoder such that $\mW^i = (\mW^{n-i})^T$, where $n$ is the total number of layers.
Literature refers to autoencoders with this type of encoder-decoder relation as weight-tied.

The convolutional autoencoder keeps the overall structure of the traditional autoencoder but replaces the fully connected layers with convolutions:
\begin{equation}
	\tL^{i+1} = \sigma(\tW^i\ast\tL^i + \vb^i),
\end{equation}
where $\ast$ denotes the convolution operation and the bias $\vb^i$ is broadcast to match the shape of $\tL^i$ such that the j-th entry in $\vb^i$ is added to the j-th channel in $\tL^i$.
Whereas before the hidden code was an m-dimensional vector, it is now a tensor with a rank equal to the rank of the input tensor.
In the case of images, that rank is three (height, width, and the number of channels).
CAEs generally include pooling layers or convolutions with strides $>$ 1 or dilation $>$ 1 in the encoder to reduce the size of the input.
In the decoder, unpooling or transposed convolution layers \citep{dumoulin16} inflate the latent code to the size of the input.
\subsection{Our Model} \label{model}
Our model consists of five strided convolution layers in the encoder and five up-sampling convolution layers (bilinear up-sampling followed by padded convolution) \citep{odena16} in the decoder.
We chose to use five layers so that the size of the latent code, after the strided convolutions, would be 4x4 or 3x3 depending on the dataset.
To increase the level of abstraction in the latent code, we increased the depth of the network by placing two residual blocks \citep{he16} with two convolutions each after each strided / up-sampling convolution layer.
We applied instance normalization \citep{ulyanov16} and ReLU activation \citep{nair10} following every convolution in the architecture.

Our goal was to understand the effect latent code shape has on different aspects of the network.
Therefore, we wanted to change the shape of the bottleneck between experiments while keeping the rest of the network constant.
To this end, we quadrupled the number of channels with every strided convolution $s^i$ and reduced it by a factor of four with every up-sampling convolution $u^i$.
In effect, this means that the volume (i. e., height $\times$ width $\times$ channels) of the feature maps is identical to the input in all layers up to the bottleneck:
\begin{align}
		s^i(\tL^i) \in \R^{\rfrac{h^i}{2} \times \rfrac{w^i}{2} \times 4n_c^i} & \text{ , for  } \tL^i \in \R^{h^i \times w^i \times n_c^i} \\	
		u^i(\tL^i) \in \R^{2h^i \times 2w^i \times \rfrac{n_c^i}{4}} & \text{ , for  } \tL^i \in \R^{h^i \times w^i \times n_c^i}
\end{align}
In this regard, our model, differs from CAEs commonly found in literature, where it is customary to double/halve the number of channels with every down-/up-sampling layer.
However, our scheme allows us to test architectures with different bottleneck shapes while ensuring that the volume of the feature maps stays the same as the input until the bottleneck. In this sense, the bottleneck is the only moving part in our experiments. 
The resulting models have a number of parameters ranging from $\sim$ 50M to 90M, depending on the bottleneck shape.
\subsection{Datasets} \label{datasets}
To increase the robustness of our study, we conducted experiments on three different datasets.
Additionally, the three datasets allowed us to address the question, how the difficulty of the data (i. e., the complexity of the data distribution) affects learning in the CAE.
To study this effect, we decided to run our experiments on three datasets of varying difficulty.
We determined the difficulty of each dataset based on intuitive heuristics.
In the following, we present the datasets in the order of increasing difficulty and our reasoning for the difficulty grading.
\subsubsection{Pokemon}
The first dataset is a blend of the images from ``Pokemon Images Dataset''\footnote{\href{https://www.kaggle.com/kvpratama/pokemon-images-dataset}{https://www.kaggle.com/kvpratama/pokemon-images-dataset}} and the type information from ``The Complete Pokemon Dataset''\footnote{\href{https://www.kaggle.com/rounakbanik/pokemon}{https://www.kaggle.com/rounakbanik/pokemon}}, both of which are available on Kaggle.
Our combined dataset consists of 793 256$\times$256 pixel images of Pokemon and their primary and secondary types as labels.
To keep the training time within acceptable bounds, we resized all images to be 128$\times$128 pixels.
We chose this dataset primarily for its clear structure and simplicity.
The images depict only the Pokemon without background, and each image centers on the Pokemon it is showing.
Additionally, the color palette in the images is limited, and each image contains large regions of uniform color.
Due to the above reasons, we deemed this dataset to be the ``easy'' dataset in our experiments.
\subsubsection{CelebA}
A step up from the Pokemon dataset in terms of difficulty is the CelebA faces dataset \citep{liu15}.
This dataset is a collection 202,600 images showing celebrity faces, each with a 40-dimensional attribute vector (attributes such as smiling/not smiling, male/female).
Since the images also contain backgrounds of varying complexity, we argue that this leads to a more complex data distribution.
Furthermore, the lighting conditions, quality, and facial orientation can vary significantly in the images.
However, some clear structure is still present in this dataset, as the most substantial portion of each image shows a human face.
For those reasons, we defined this dataset to have ``medium'' difficulty.
For our purposes, we resized the images to be 96$\times$96 pixels.
The original size was 178$\times$218 pixels.
\subsubsection{STL-10}
For our last dataset, we picked STL-10 \citep{coates11}. 
This dataset consists of 96$\times$96 pixel natural images and is divided into three splits: 5,000 training images (10 classes), 8,000 test images (10 classes), 100,000 unlabeled images.
The unlabeled images also include objects that are not covered by the ten classes in the training and test splits.
As the images in this dataset show many different scenes, from varying viewpoints and under a multitude of lighting conditions, we find samples from this dataset to be the most complex and, therefore, the most difficult of the three.
\section{Experiments} \label{experiments}
\subsection{Autoencoder Training} \label{training}
The first experiment we conducted, and which forms the basis for all subsequent experiments, consists of training of autoencoders with varying bottleneck sizes and observing the dynamics of their training and test losses.
This experiment probes the relative importance of latent code size versus its number of channels.
Additionally, we mean to provide insight into how and when our models overfit and if the data complexity (see Section \ref{datasets}) plays a discernible role in this.
We also tested the widespread hypothesis that autoencoders learn to ``copy'' the input if there is no bottleneck.
On the Pokemon dataset, we trained the CAEs on the first 634 images and reserved the remaining 159 for testing.
To be able to observe overfitting behavior, we used only the first 10,000 images in the CelebA dataset for training and the last 2,000 images for testing.
For STL-10 we used images from the unlabeled split, as they are not limited to the 10 classes in either the training or test splits.
Analogously to CelebA, we used the first 10,000 images for training and the last 2,000 for testing of the CAEs.

For each dataset, we selected three latent code sizes (=height=width) $s_i$, $i \in \{1, 2, 3\}$ as
\begin{equation}
	s_i = \frac{s_{input}}{2^{n_l-i+1}} \text{\hspace{15pt} }i \in \{1, 2, 3\},\hspace{5pt} n_l=5
\end{equation}
In this equation, $n_l = 5$ is the number of strided convolutions in the network, and $s_{input}$ is the height (= width) of the images in the dataset.
Throughout the rest of the paper, we mean width and height when we refer to the size of the bottleneck.
To obtain latent codes with size $s_2$ ($s_3$), we changed the strides in the last (two) strided convolution layer(s) from two to one.
For each size we then fixed four levels of compression $c_j \in \{\rfrac{1}{64}, \rfrac{1}{16}, \rfrac{1}{4}, 1\}$ and calculated the necessary number of channels $n_{c_j}$ according to
\begin{equation}
	n_{c_j} = \frac{c_j s_{in}^2 n_{c_{in}}}{s_i^2}\text{\hspace{15pt} }i \in \{1, 2, 3\},\hspace{5pt} j \in \{1,2,3,4\}
\end{equation}
Here, $n_{c_{in}}$ is the number of channels in the input image.
This way, the autoencoders had the same number of parameters in all layers except the ones directly preceding and following the bottleneck.
We used mean squared error (MSE) between reconstruction and input as our loss function.
After initializing all models with the same seed, we trained each for 1,000 epochs, computing the test error after every epoch.
We repeated this process for three different seeds.
\subsection{Knowledge Transfer} \label{classifiers}
Another goal of our investigation was to estimate the effect of the latent code shape on transferability.
Here, our idea was to train a logistic regression on latent codes to predict the corresponding labels for each dataset.
Since logistic regression can only learn linear decision boundaries, this approach allows us to catch a glimpse of the sort of knowledge present in the latent code and its linear separability.
Furthermore, this serves as another test for the ``copying'' hypothesis.
If the encoder has indeed learned to copy the input, the results of the logistic regression will be the same for the latent codes and the input images.
We performed this experiment separately for data which the CAEs saw during training and unseen data.
In the first step, we exported latent codes for all 108 trained models.
In the case of the Pokemon dataset, we exported the full dataset.
From the CelebA dataset, we used the 10,000 samples in the training data and another 10,000 unseen images from the end of the dataset.
Since we trained CAEs on samples from the unlabeled split of STL-10, there were no labels available for the training data.
Consequently, we only used the 8,000 images in the STL-10 test split for classification but also extracted the representations of the 10,000 training samples for another experiment (see Section \ref{svcca}).

After initial attempts, we noticed that a single learning rate and training duration is not adequate for training all models, most likely due to variation in the dimensionality of the representations.
To ensure that our comparison is as fair as possible, we decided to apply a uniform heuristic instead of a single learning rate and training duration.
We used the fast.ai framework to train the classifiers with a one-cycle policy \citep{smith2018} and early stopping for a maximum of 100 epochs.
For each classifier, we used 60\% of data for training, 20\% as a validation set for early stopping, and 20\% for final testing.
Additionally, each classifier was trained three times on different seeds and different samples in the splits.
Besides, we also trained models directly on the image data for every dataset to serve as a baseline for comparison (also with three seeds).
\subsection{Scaling with Dataset Size} \label{scaling}
In order to be able to observe overfitting in the training of the CAEs, we limited the training data of the CelebA and STL-10 datasets to 10,000 samples.
To estimate what effect the amount of training data had on CAE training, we repeated a toned-down version of our first experiment for six different amounts of training data.
For this experiment, we chose to use the CelebA dataset and four of the twelve models from the first experiment (3$\times$3$\times$48, 3$\times$3$\times$192, 6$\times$6$\times$12, 6$\times$6$\times$48).
We trained the models on the first 1\% (2,000 samples), 5\% (10,000 samples), 10\% (20,000 samples), 25\% (50,000 samples), 50\% (100,000 samples), and 95\% (190,000 samples) of the samples in the dataset while reserving the last 5\% for testing.
Each model trained for 35,000 iterations with a batch size of 128, which translates to roughly half of the compute each model received in the first experiment.
For each split and model, we used three random seeds, which were different from the ones used in Section \ref{training}.
\subsection{Pair-wise Representation Similarity} \label{svcca}
In our final experiment, we used the singular vector canonical correlation analysis (SVCCA) \citep{raghu17} technique to gauge the pair-wise similarity of the learned latent codes.
SVCCA takes two sets of neuron activations of the shape number of neurons $\times$ data points and estimates aligned directions in both spaces that have maximum correlation.
First, SVCCA calculates the top singular vectors that explain 99\% of the variance using singular value decomposition (SVD).
Subsequently, SVCCA finds affine transformations for each set of singular vectors that maximize their alignment in the form of correlation.
Lastly, it averages the correlation for each direction in the discovered subspace to produce a scalar similarity score.
In convolutional neural networks, this computation can become prohibitively expensive, due to the large size of the feature maps.
For such cases, \citet{raghu17} recommends transforming the feature maps using discrete Fourier transformation (DFT).
In the publication, the authors show that DFT leaves SVCCA invariant (if the dataset is translation invariant) but results in a block diagonal matrix, which enables exact SVCCA computation by computing SVCCA for each neuron at a time.
Additionally, they recommend down-sampling bigger feature maps in Fourier space when comparing them to smaller ones.
In this experiment, we investigated the effect of latent code shape on its structure and content.
\section{Results and Discussion} \label{results}
\captionsetup[subfigure]{labelformat=empty}
\begin{figure}[ht]
	\centering
	\includegraphics[width=\linewidth]{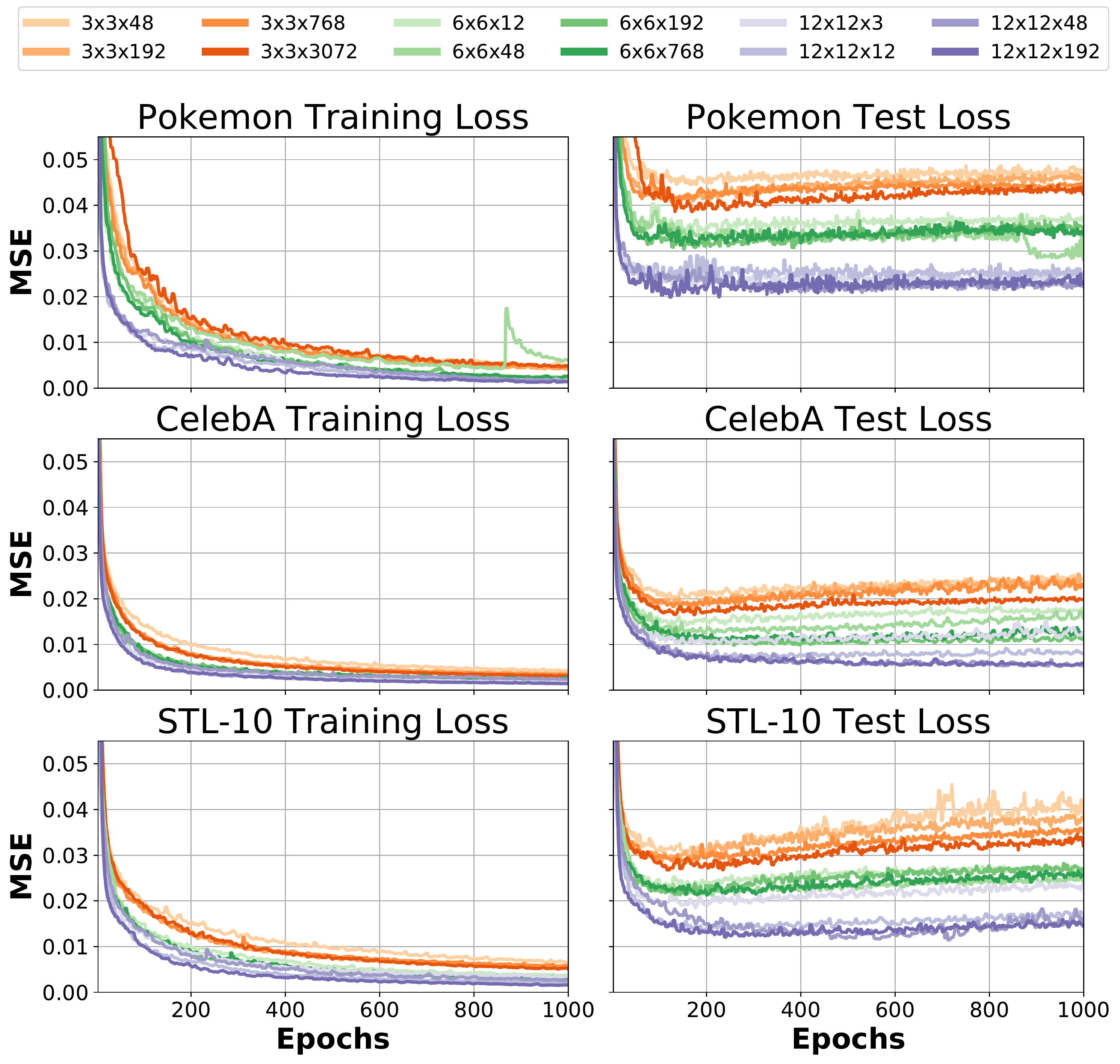}
	\caption{Loss plots for the three datasets averaged over the three seeds. Each row corresponds to a dataset. From top to bottom: Pokemon, CelebA, STL-10. The left column shows the training error, while the right one depicts test error. Every bottleneck configuration is shown as a distinct line. Configurations that have a common feature map size share the same color. Color intensity represents the amount of channels in the bottleneck (darker = more channels). In the case of Pokemon, the feature map sizes are 4, 8, 16 instead of 3, 6, 12} \label{losses}
\end{figure}

\captionsetup[subfigure]{labelformat=parens, position=bottom}
\begin{figure}[t]
	\centering
	\subfloat[Pokemon Training]{\includegraphics[width=0.48\linewidth]{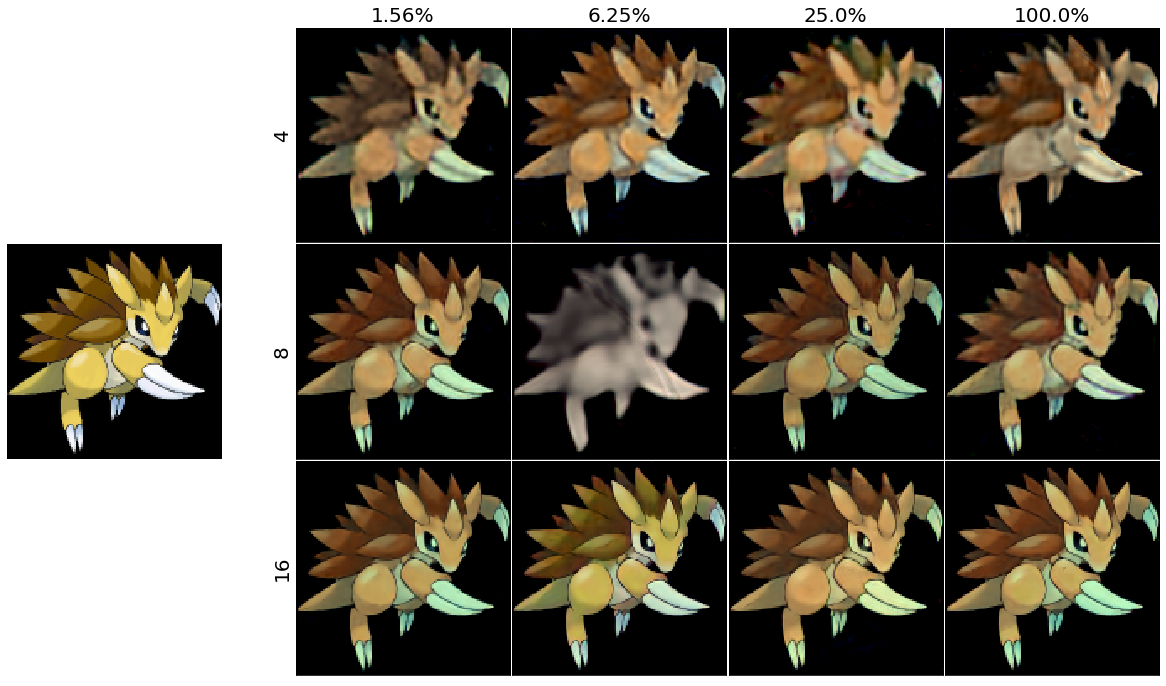}}\hfill
	\subfloat[Pokemon Test]{\includegraphics[width=0.48\linewidth]{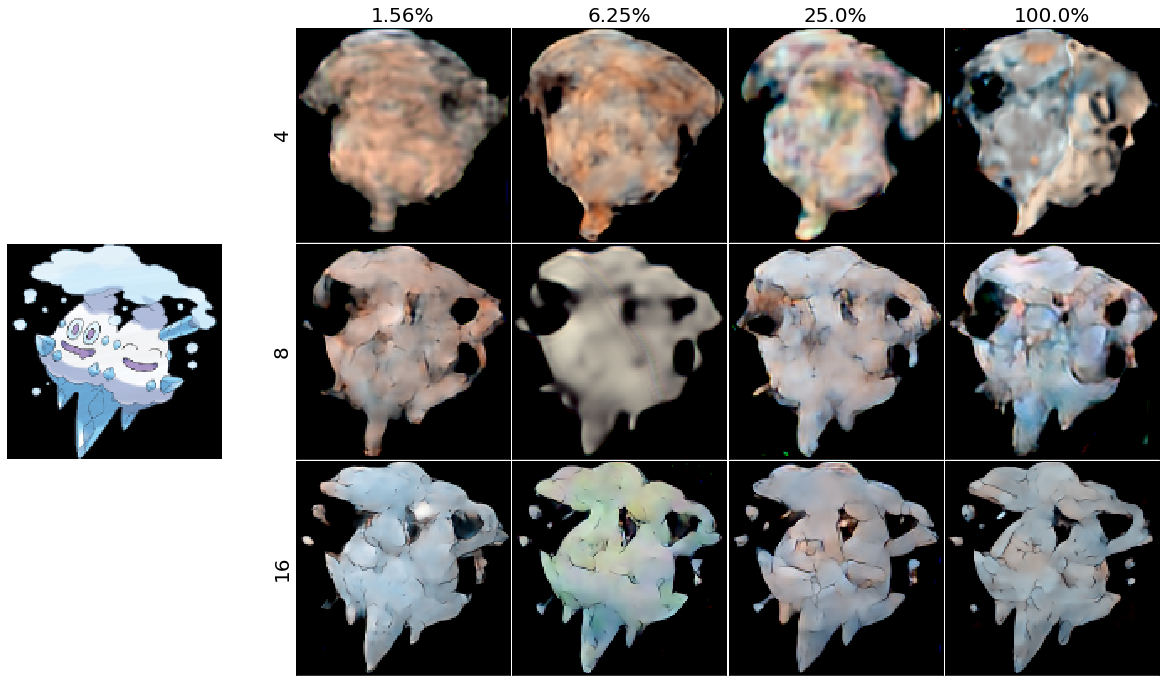}}\\
	\subfloat[CelebA Training]{\includegraphics[width=0.48\linewidth]{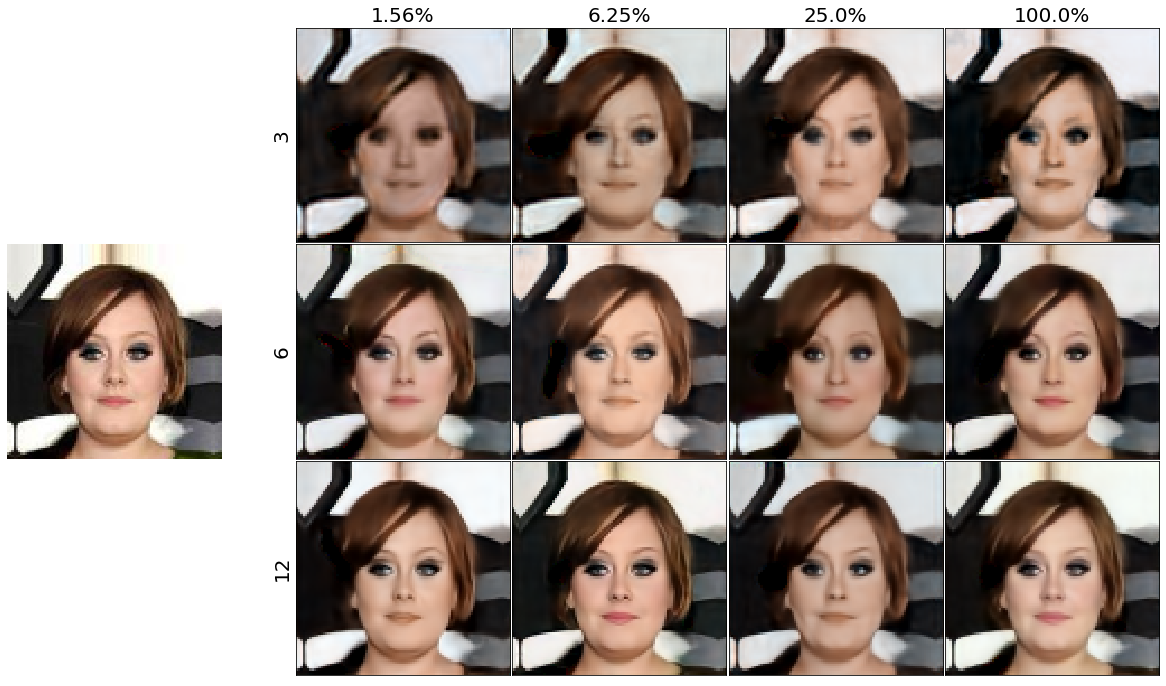}}\hfill
	\subfloat[CelebA Test]{\includegraphics[width=0.48\linewidth]{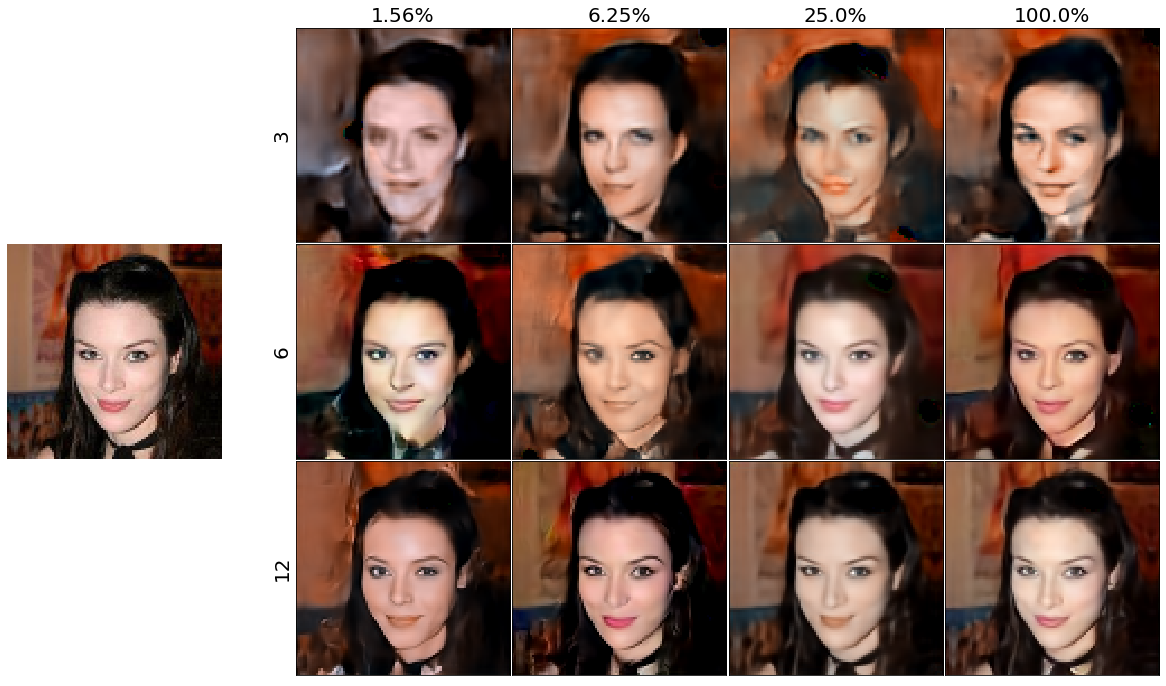}}\\
	\subfloat[STL-10 Training]{\includegraphics[width=0.48\linewidth]{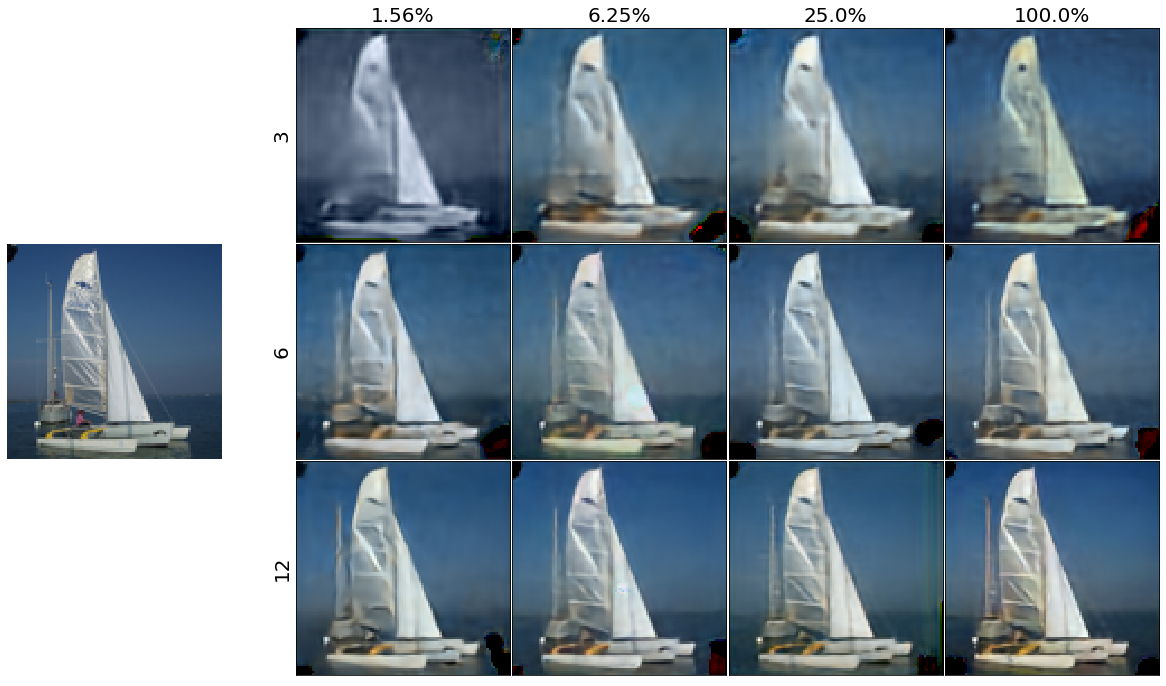}}\hfill
	\subfloat[STL-10 Test]{\includegraphics[width=0.48\linewidth]{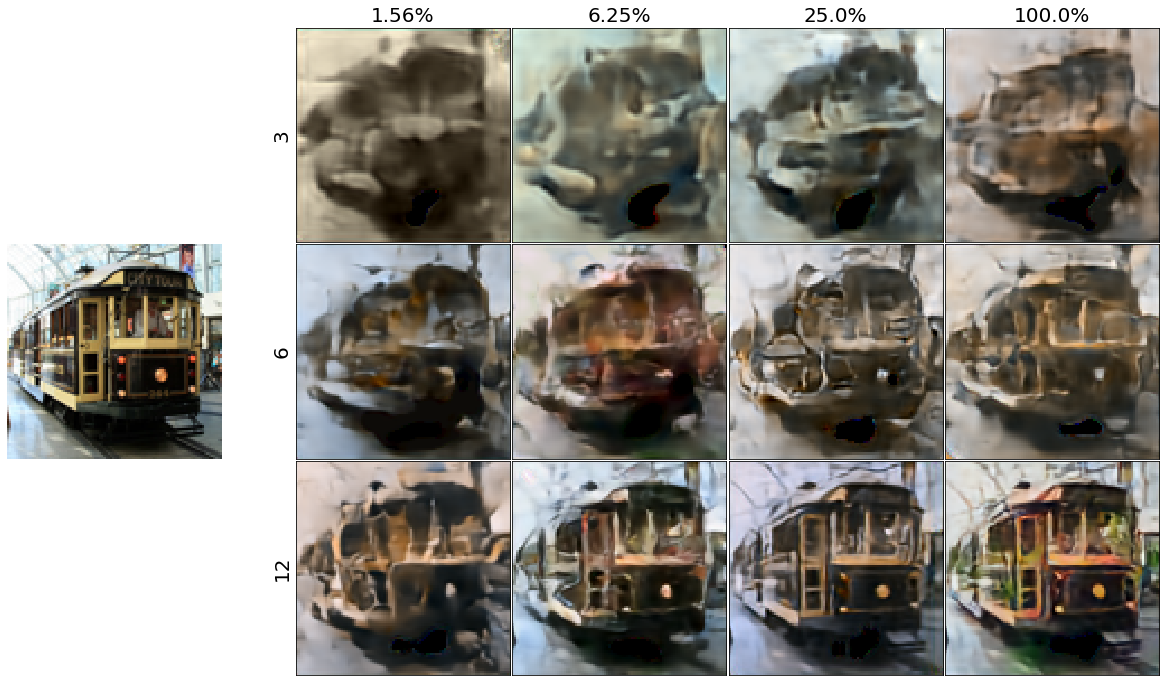}}
	\caption{Reconstructions of randomly picked samples for the models initialized with seed 0. The left column contains reconstructions from the training data, while on the right, we show reconstructions from unseen samples. In each subfigure, the rows correspond to CAEs with the same bottleneck size (height, width), increasing from top to bottom. The columns group CAEs by the number of channels in the bottleneck, expressed as percentage relative to input given bottleneck size. The image to the left of each grid is the original input.} \label{samples}
\end{figure}

Looking at the error curves for the CAEs (Fig. \ref{losses}), we make several observations:
\begin{enumerate}
	\item{The total amount of neurons in the bottleneck does not affect training as much as expected.
	All CAEs converge to a similar training error.
	We find this surprising, as the smallest bottlenecks have only 1.56\% of total neurons compared to the largest ones.
	Although the final differences in training error are small, we discover that the size of the bottleneck feature maps has a more substantial effect on training error than the number of channels.
	The larger the bottleneck width and height, the lower the training error.
	This correlation is particularly visible in the earlier stages of training, indicating that CAEs with a bigger bottleneck size are faster to train.
	An interesting outlier presents itself in the plots for the Pokemon dataset.
	Here, we see that late in the training of the CAE with the 8x8x48 bottleneck training error suddenly spikes.
	At the same time, the test error drops significantly approximately to the same level as the training error.
	We verified that this was not due to an unintended interruption in training, by retraining the model with the same seed and obtained an identical result.
	Out of the 108 CAEs we trained for the first experiment and the 72 CAEs for the third, we observed this kind of behavior only once.
	Currently, it is unclear to us how such a drastic change in model parameters came about at such a late stage in training.
	Usually, we expect the loss landscape to become smoother the longer we train a model \citep{goodfellow14}.
	Whether this outlier is a fluke or has implications for the loss landscape of CAEs remains to be seen as our understanding of the training dynamics of neural networks deepens.}
	\item{We observe that bottleneck shape critically affects generalization.
	Increasing the number of channels in the bottleneck layer seems to improve test error only slightly and not in all cases.
	The relationship between bottleneck size and test error, on the other hand, is clear cut.
	Larger bottleneck size correlates with a significant decrease in test error.
	This finding is surprising, given the hypothesis that only the total amount of neurons matters.
	The CAE reconstructions further confirm this hypothesis.
	We visually inspected the reconstructions of our models (samples are shown in Fig. \ref{samples} and in the supplementary material) and found that reconstruction quality improves drastically with the size of the bottleneck, yet no so much with the number of channels.
	As expected from the loss plots, the effect is more pronounced for samples from the test data.}
	\item{Bottleneck shape also affects overfitting dynamics.
	We would expect the test score to increase after reaching a minimum, as the CAE overfits the data.
	Indeed, we observe this behavior in some cases, especially in CAEs with smaller bottleneck sizes or the minimum amount of channels.
	In other cases, predominantly in CAEs with a larger bottleneck size, the test error appears to plateau instead.
	In the plot for the CelebA dataset, the curves for 12x12x48 and 12x12x192 even appear to decrease slightly over the full training duration.
	This overfitting behavior implies that CAEs with a larger bottleneck size can be trained longer before overfitting occurs.}
	\item{CAEs, where the total number of neurons in the bottleneck is the same as the number of pixels in the input, do not show signs of simply copying images.
	If the CAEs would indeed copy images, the test error would go to zero, yet we do not observe this case in any of the datasets.
	What is more, these complete CAEs follow the same pattern as the under-complete ones and often converge to similar values.
	This finding directly contradicts the popular hypothesis about copying CAEs.
	In essence, it suggests that even complete CAEs learn abstractions from data, and raises the question: What prevents the CAE from simply copying its input?
	We believe that the answer to this question could potentially lead to new autoencoder designs that exploit this limitation to learn better representations.
	Hence, we argue that it is an exciting direction for future research.
	Additionally, the trends we derive from our results suggest that this finding likely extends to over-complete CAEs as well.
	However, experiments with over-complete CAEs are required to test this intuition.}
\end{enumerate}
\begin{figure}[t!]
	\newcommand{\subwidth}{0.48\linewidth}
	\centering
	\subfloat{\includegraphics[width=\subwidth]{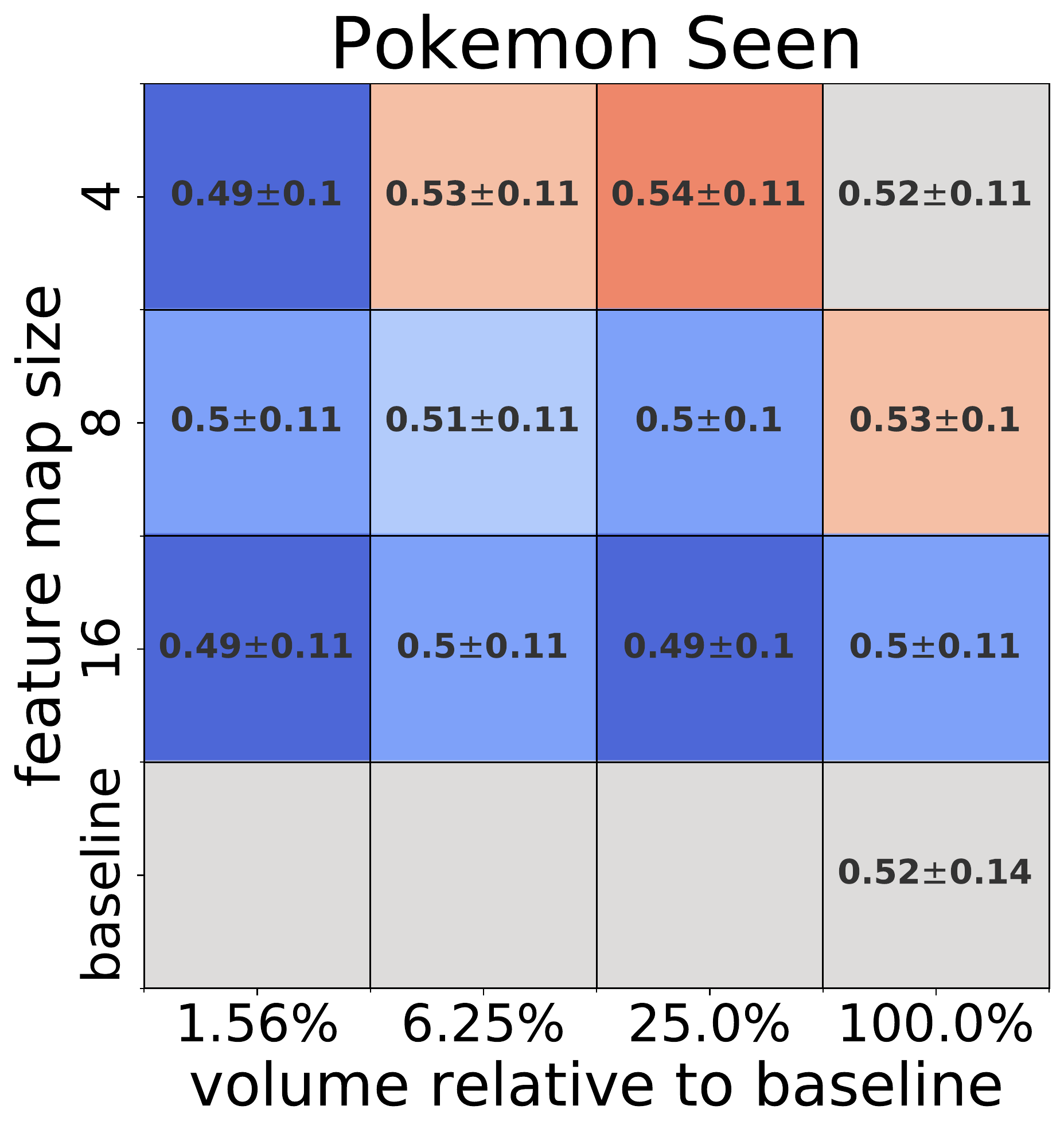}}\hfill\vspace{-10pt}
	\subfloat{\includegraphics[width=\subwidth]{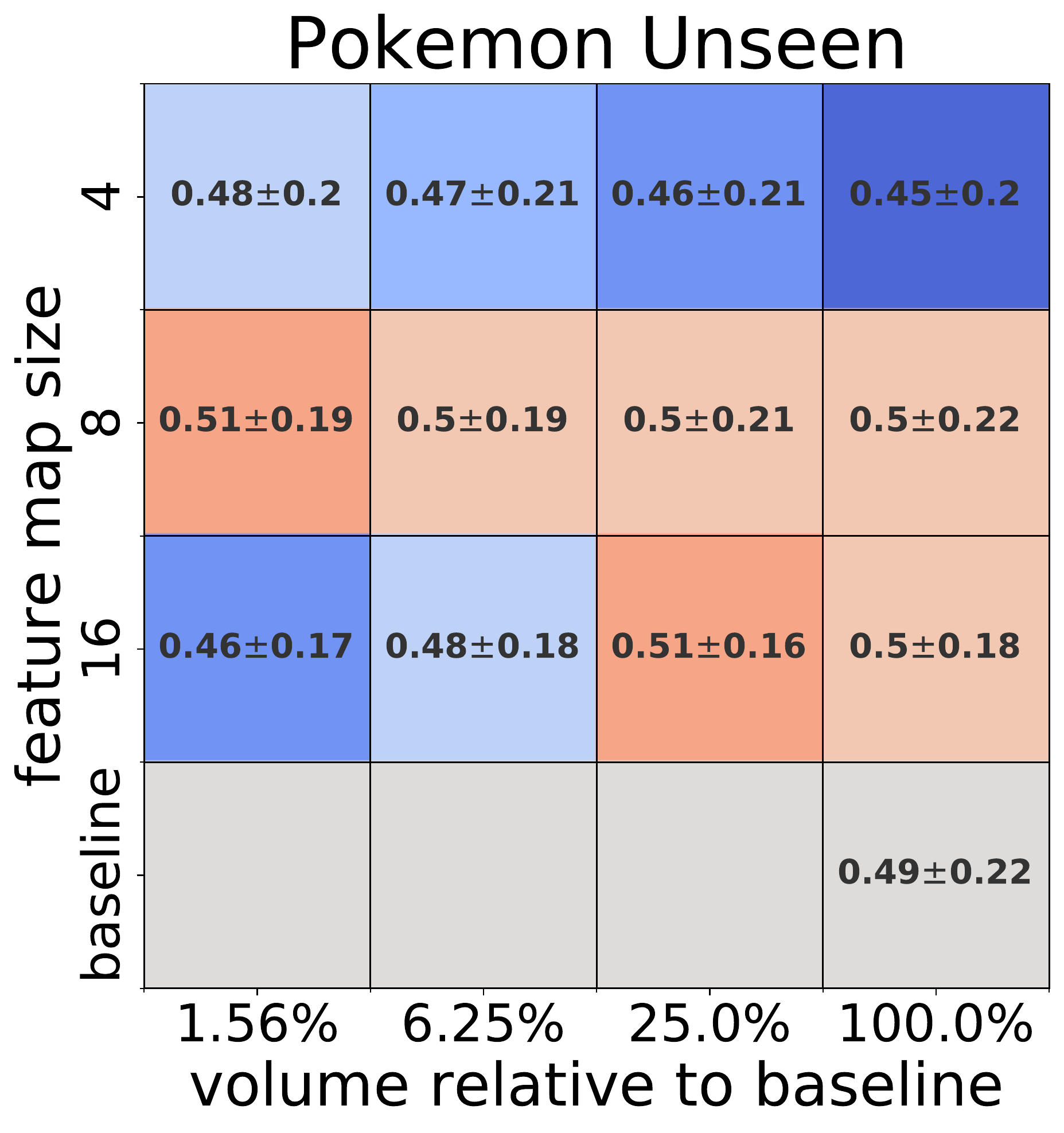}} \\
	\subfloat{\includegraphics[width=\subwidth]{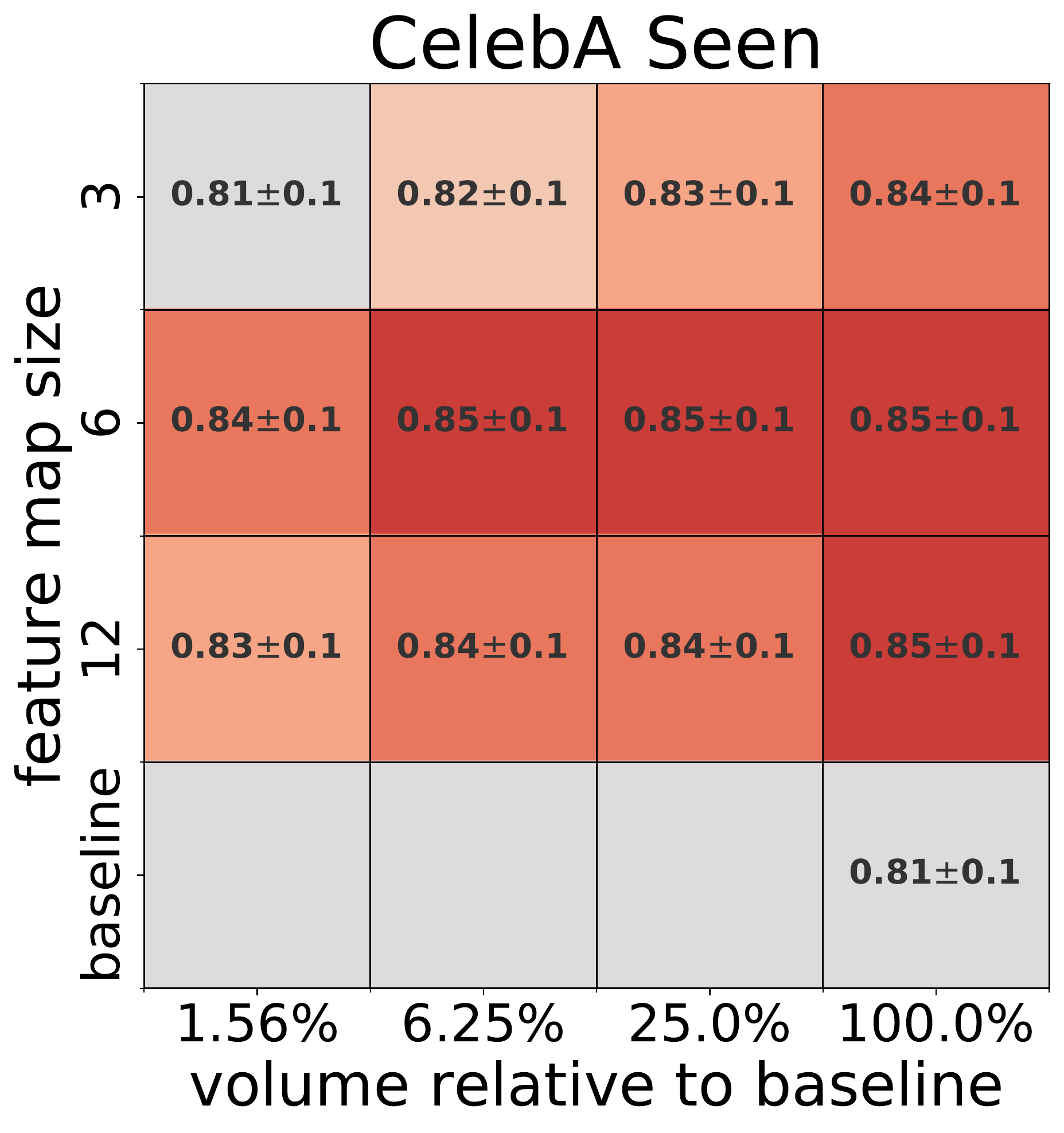}}\hfill\vspace{-10pt}
	\subfloat{\includegraphics[width=\subwidth]{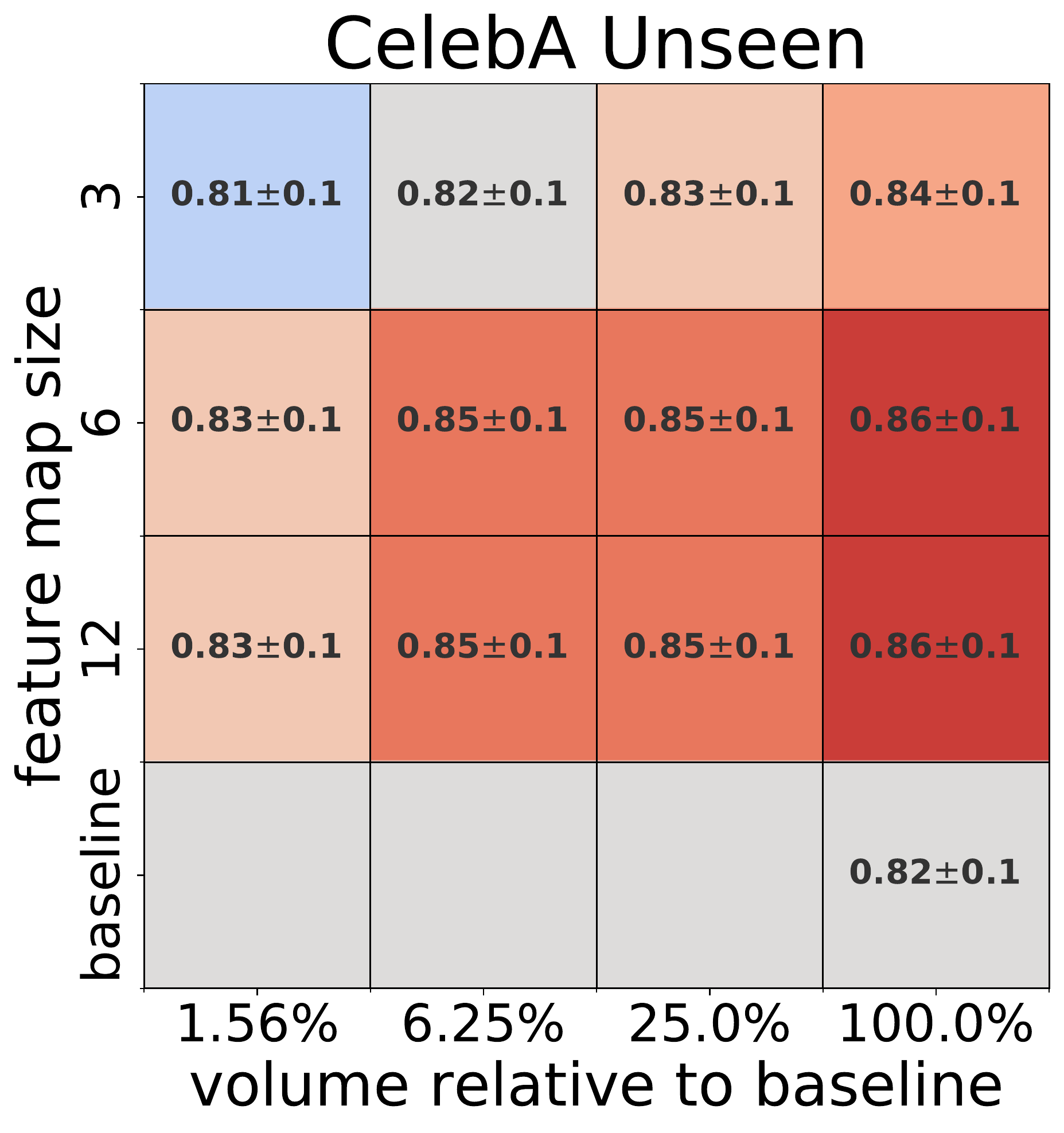}} \\
	\subfloat{\includegraphics[width=\subwidth]{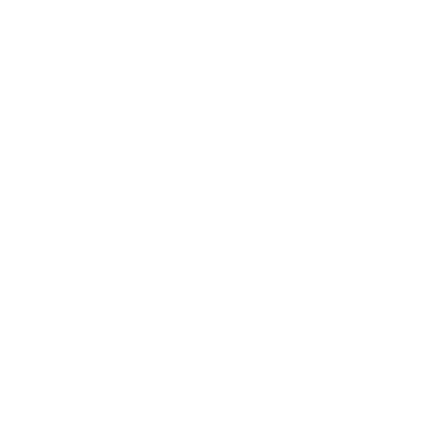}}\hfill
	\subfloat{\includegraphics[width=\subwidth]{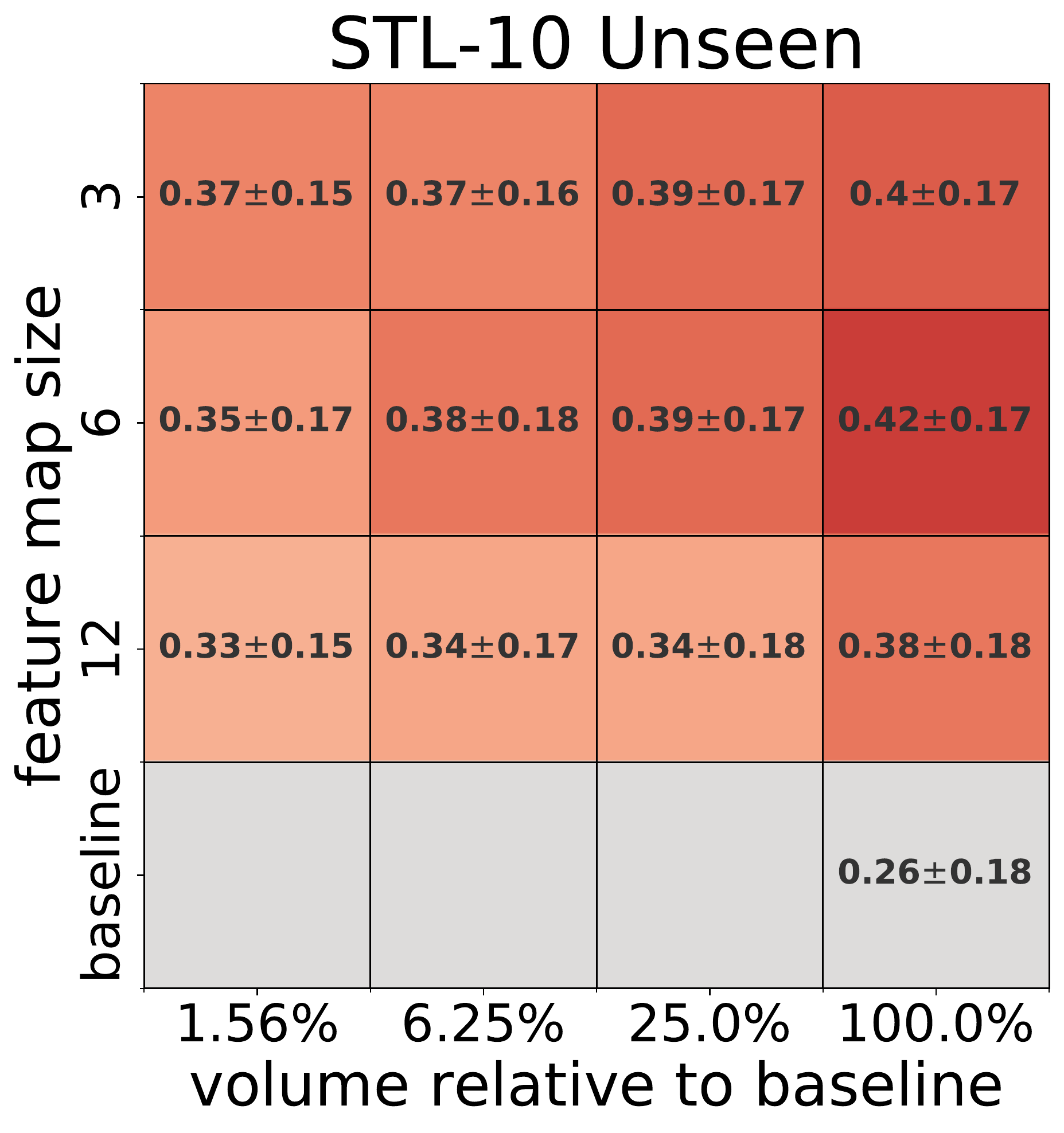}}
	\caption{Results from training linear models on latent codes to predict the labels associated with each dataset. For Pokemon and CelebA ROC AUC is shown. The plots for STL-10 show macro F1 score. The left column corresponds to classification of samples that the CAE has seen during training, while the right column is from unseen samples. Color is based on difference to baseline (i.e., linear model trained directly on images), where red signifies an improvement.} \label{classifier-plots}
\end{figure}

Furthermore, the loss curves and reconstruction samples appear to only marginally reflect the notion of dataset difficulty, as defined in Section \ref{datasets}.
One thing that stands out is the large generalization gap on the Pokemon dataset, which is most likely due to the comparatively tiny dataset size of $\approx$ 600 training images.
Comparing the results for CelebA and STL-10, we find that overall generalization appears to be slightly better for CelebA, which is the less difficult dataset of the two.
The test errors on STL-10 exhibit higher variance than on CelebA, although the number of samples and training epochs are equal between the two.
This effect also shows itself in the reconstruction quality.
On CelebA, even the CAEs with the smallest bottlenecks manage to produce decent reconstructions on test data, whereas reconstructions on STL-10 are often unrecognizable for the corresponding models.
Overall, this effect is not clear-cut and warrants a more thorough investigation of the relationship between data complexity and CAE characteristics, especially in the light of compelling results from curriculum learning research \citep{bengio09}.
\begin{figure}[t]
	\centering
	\includegraphics[width=\linewidth]{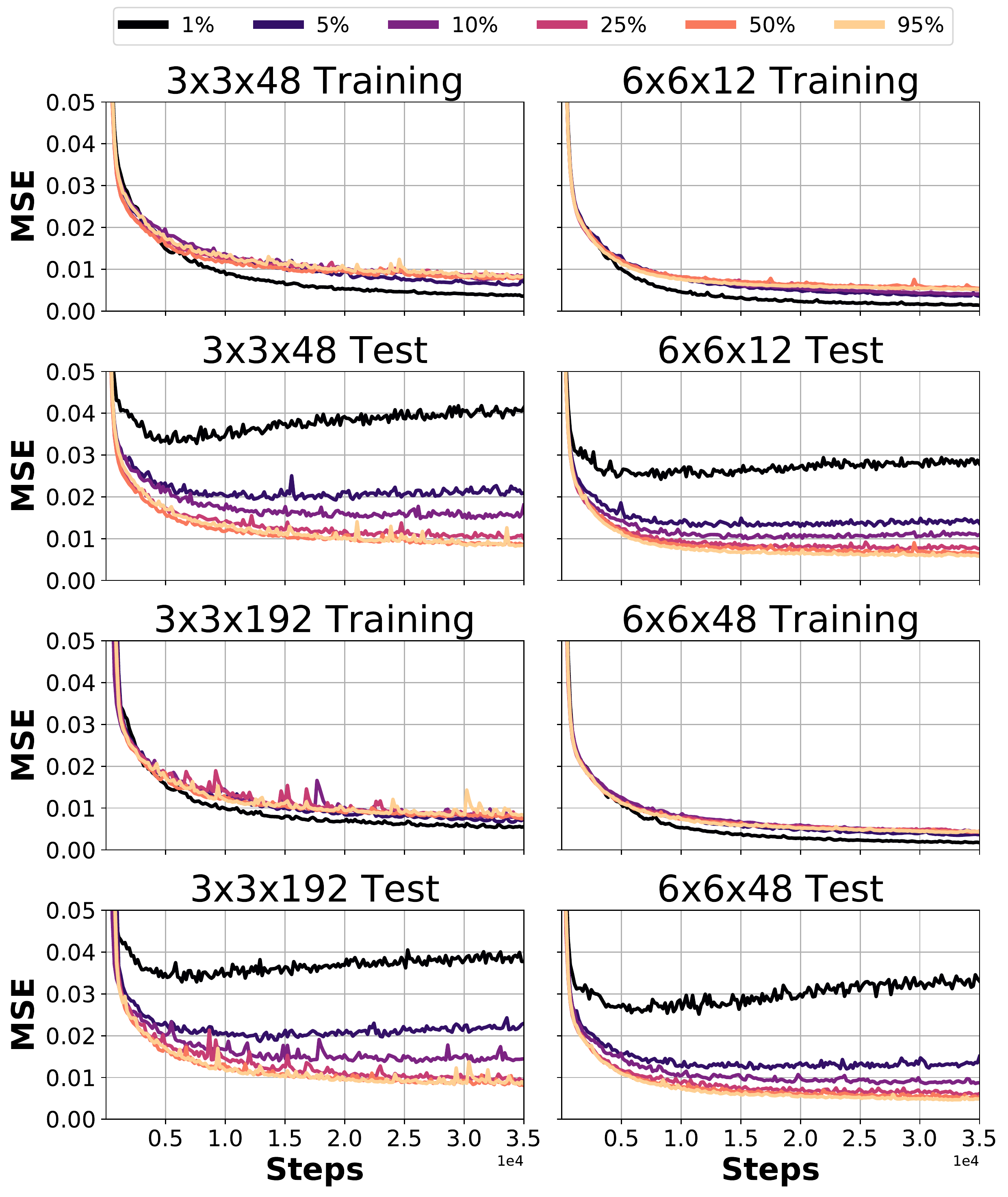}
	\caption{Loss plots for the scaling experiment (see Section \ref{scaling}) averaged over three runs. Each quadrant shows training and test losses from one model. For better comparison, we juxtaposed models with the same number of neurons in the bottleneck. Each color corresponds to the fraction of the data the model was trained on, as shown in the legend at the top.} \label{scaling-losses}
\end{figure}

If we look at the results of our knowledge transfer experiments (Fig. \ref{classifier-plots}), we find further evidence that contradicts the copying autoencoder hypothesis.
Although the loss curves and reconstructions already indicate that the CAE does not copy its input, the possibility remains that the encoder distributes the input pixels along the channels but the decoder is unable to reassemble the image.
Here, we see that the results from the linear model trained on latent codes perform better than the ones trained on the inputs (marked ``baseline'' in the figure).
The only deviation from this pattern seems to be the Pokemon dataset, where the average performance is more or less the same for all settings.
However, even in this case, we find that CAE representations and baseline perform differently on individual classes (see supplementary material).
As such, it seems implausible to assume that the encoder copied the input to the bottleneck.
Overall, we find that knowledge transfer also seems to work slightly better on latent codes with greater size, although the effect is relatively weak.
Interestingly, the number of channels in the code seems to improve the performance of the classification.
It is not clear, however, whether this effect is due to the structure of the codes or their high dimensionality.
Perhaps projecting the representations to have the same dimensionality (using PCA or UMAP) before classification could answer this question.
Additionally, we find that the performance of the classifier is almost the same on seen and unseen data.
Given the differences in overfitting between the different models, we would expect otherwise.
This may suggest that encoder and decoder exhibit different degrees of overfitting.

\begin{figure}[t]
	\centering
	\includegraphics[width=\linewidth]{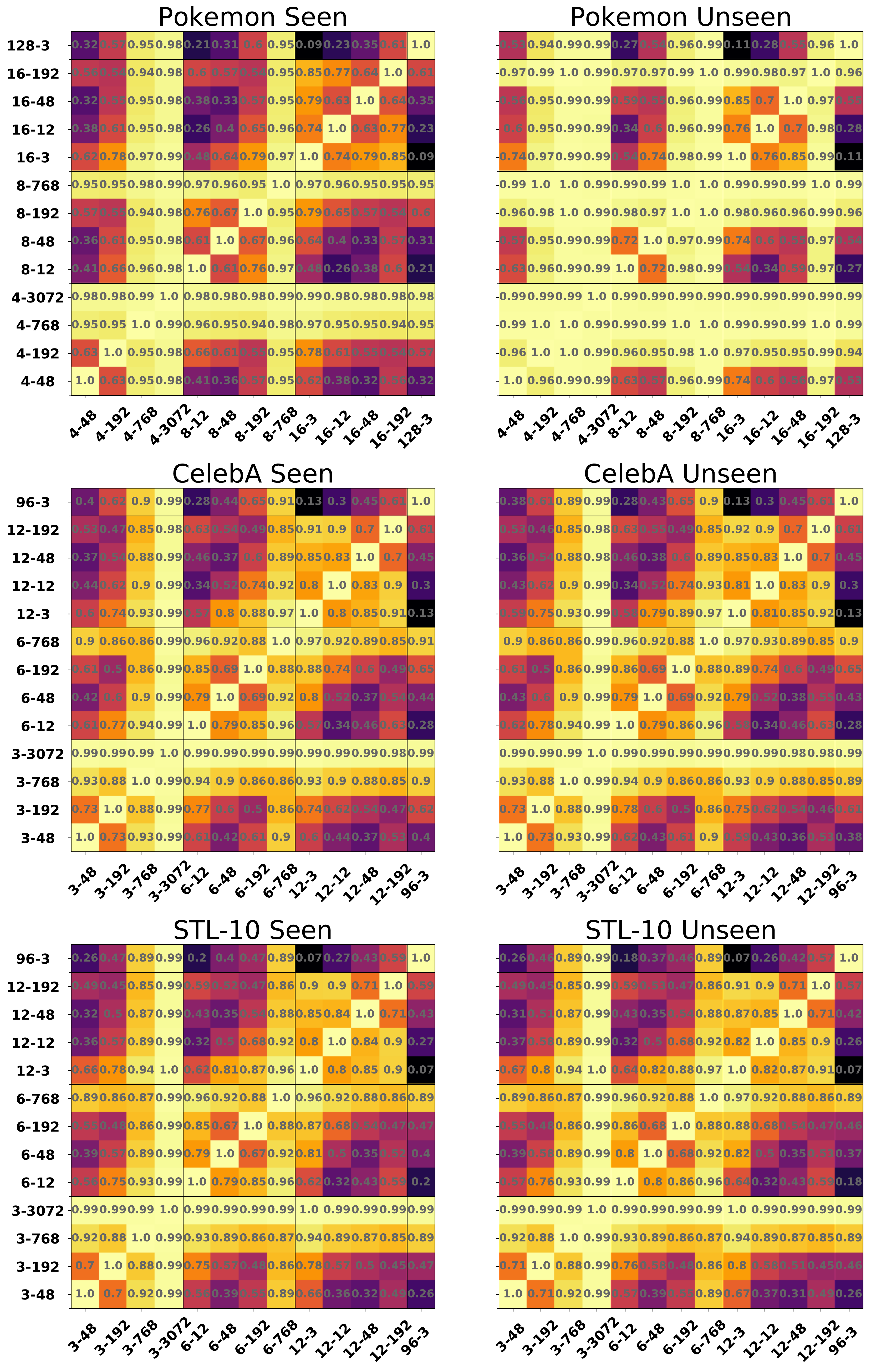}
	\caption{Results of pair-wise SVCCA. Labels on the x and y axis correspond to (height=width)-(number of channels) in the bottleneck.} \label{svcca-plots}
\end{figure}
Another point of interest to us is that the classification of training or test samples of the CAE does not result in an appreciable difference in performance.
From the evident overfitting of the CAEs as seen in the reconstructions and loss curves, we expected otherwise.
This discrepancy raises the question if overfitting happens mostly in the decoder, while the encoder retains most of its generality.
We believe that this question warrants further investigation, especially in light of the recent growth in the popularity of transfer learning methods.

We present the results of the dataset scaling experiment from Section \ref{scaling} in Fig. \ref{scaling-losses}.
This figure provides several insights:
\begin{enumerate}
	\item{Training with more data unsurprisingly results in better generalization but with diminishing returns.
	When using more than 25\% of the full datasets ($\approx$ 50,000 samples), the test error does not improve significantly, as it converges to the training error.
	These models did not plateau at 35,000 steps, however, meaning that differences could arise if the models were trained longer.}
	\item{We can confirm our observation from Fig. \ref{losses} that bigger bottleneck size results in faster training.
	As can be seen from the scaling experiment, this effect is independent of training set size.
	The only exception we can see are the models trained on only 1\% of the data ($\approx$ 2,000 samples).
	Here, the models start overfitting early in training, resulting in a very low training error but poor generalization.}
	\item{The effect of the bottleneck size on generalization, which we discovered in the first experiment, is present for all fractions of the training data.
	Additionally, we find the decrease in test error to be significantly greater if we train the model on fewer samples.}
\end{enumerate}

From the results of the pair-wise similarity measurement (see Section \ref{svcca}), we notice that the latent codes from bottlenecks with the same size have higher SVCCA similarity values, as can be seen in Fig. \ref{svcca-plots} in the blocks on the diagonal.
This observation further supports our hypothesis that latent code size, and not the number of channels, dictates the tightness of the CAE bottleneck.
Finally, we wish to point out some observations in the SVCCA similarities as a possible inspiration for future research:
\begin{itemize}
	\item{Overall, similarity does not change much between seen and unseen data, except for the Pokemon dataset, where severe overfitting occurred}
	\item{Latent codes from complete CAEs show high similarity to all latent codes from all other CAEs}
	\item{SVCCA similarity with the raw inputs tends to increase with the number of channels}
\end{itemize}
\section{Conclusion}
In this paper, we presented the findings of our in-depth investigation of the CAE bottleneck.
We could not confirm the intuitive assumption that the total number of neurons sufficiently characterizes the CAE bottleneck.
We demonstrate that the height and width of the feature maps in the bottleneck are what defines its tightness, while the number of channels plays a secondary role.
Larger bottleneck size (i.e., height and width) is also critical in achieving lower training and test errors, while simultaneously speeding up training.
These insights are directly transferable to the two main areas of application for CAEs, outlier detection and compression/denoising:
In the case of outlier detection, the model should yield a high reconstruction error on out-of-distribution samples.
Using smaller bottleneck sizes to limit generalization could prove useful in this scenario.
Compression and denoising tasks, on the other hand, seek to preserve image details while reducing file size and discarding unnecessary information, respectively.
In this case, a bigger bottleneck size is preferable, as it increases reconstruction quality at the same level of compression.
Furthermore, we could not confirm the commonly held belief that complete CAE (i. e., CAEs with the same number of neurons in the bottleneck as pixels in the input) will learn to copy its input.
On the contrary, even complete CAEs appear to follow the same dynamics of bottleneck size, as stated above.
In knowledge transfer experiments, we have also shown that CAEs that overfit retain good predictive power in the latent codes, even on unseen samples.

Our investigation yielded additional results that spark new research questions.
Data complexity, as estimated by human intuition, did not lead to significant differences in the training dynamics of our models.
On the flipside, curriculum learning, which rests on a similar notion of difficulty, has been shown to lead to improvements in the training of classifiers and segmentation networks.
The link between those two empirical results is still unclear.
Another interesting question that arose from our experiments is how overfitting manifests itself in CAEs.
Does it occurs mainly in the encoder or decoder or equally in both?

\newpage
\bibliography{bibliography}
\bibliographystyle{icml2020}

\onecolumn
\newpage
\appendix
\counterwithin{figure}{section}
\section{Supplementary Material}
\newcommand{\subwidth}{0.9\linewidth}
\begin{figure}[h]
	\centering
	\includegraphics[width=\subwidth]{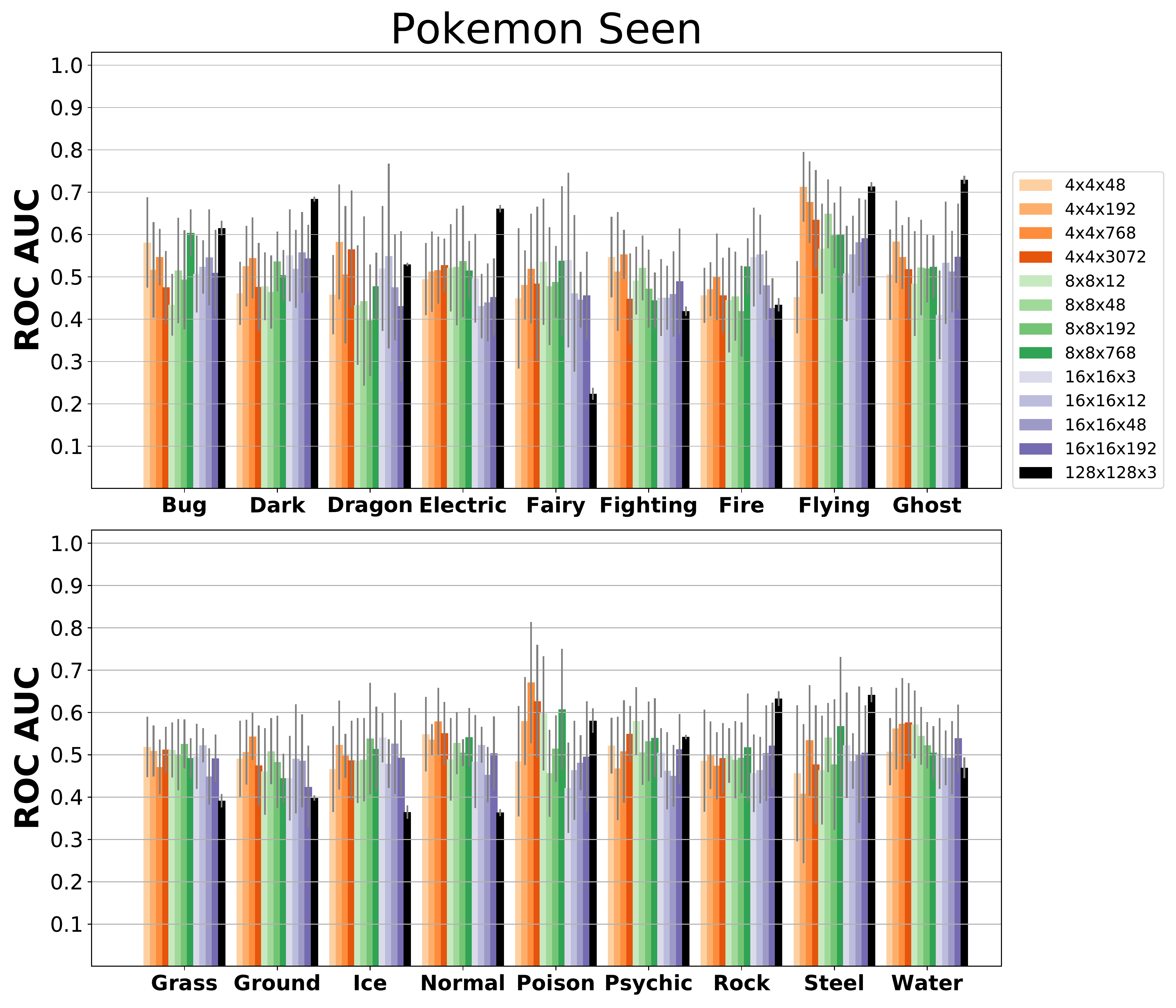}
	\caption{Classification performance by class for the Pokemon dataset. The classifiers were trained on representations from samples that the CAE has seen during training. Configurations that have a common feature map size share the same color in the bar plot. Color intensity represents the amount of channels in the bottleneck (darker = more channels). Classification based on the raw inputs (baseline) is shown in black. Error bars indicate standard deviation over 9 runs (3 CAE seeds $\times$ 3 classifier seeds), except for the baseline where only the 3 classifier seeds are considered.}
\end{figure}

\begin{figure}[h]
	\centering
	\includegraphics[width=\subwidth]{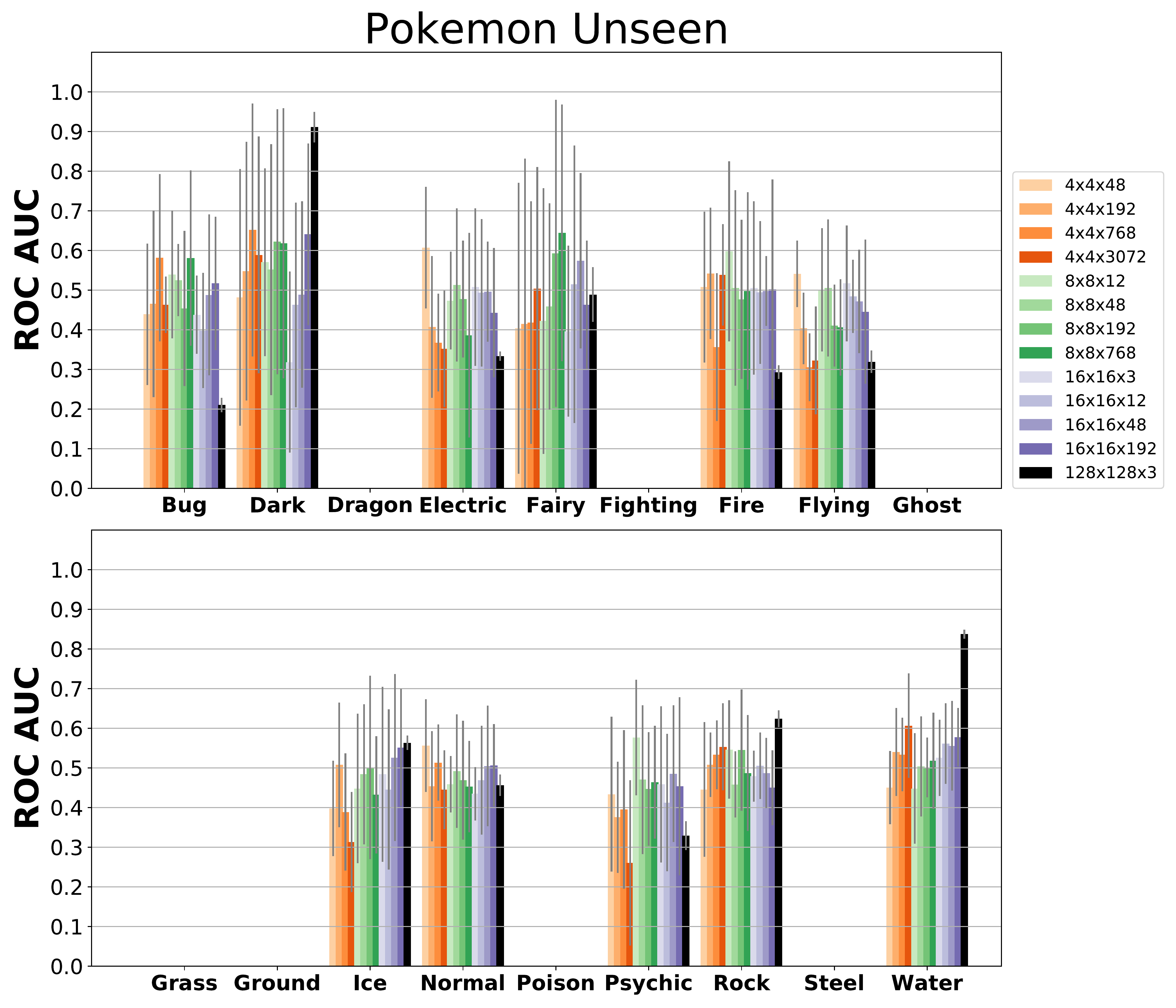}
	\caption{Classification performance by class for the Pokemon dataset. The classifiers were trained on representations from samples that the CAE has not seen during training. Configurations that have a common feature map size share the same color in the bar plot. Color intensity represents the amount of channels in the bottleneck (darker = more channels). Classification based on the raw inputs (baseline) is shown in black. Error bars indicate standard deviation over 9 runs (3 CAE seeds $\times$ 3 classifier seeds), except for the baseline where only the 3 classifier seeds are considered. Missing bars indicate that the class was not present in the data.}
\end{figure}

\begin{figure}[h]
	\centering
	\includegraphics[width=\subwidth]{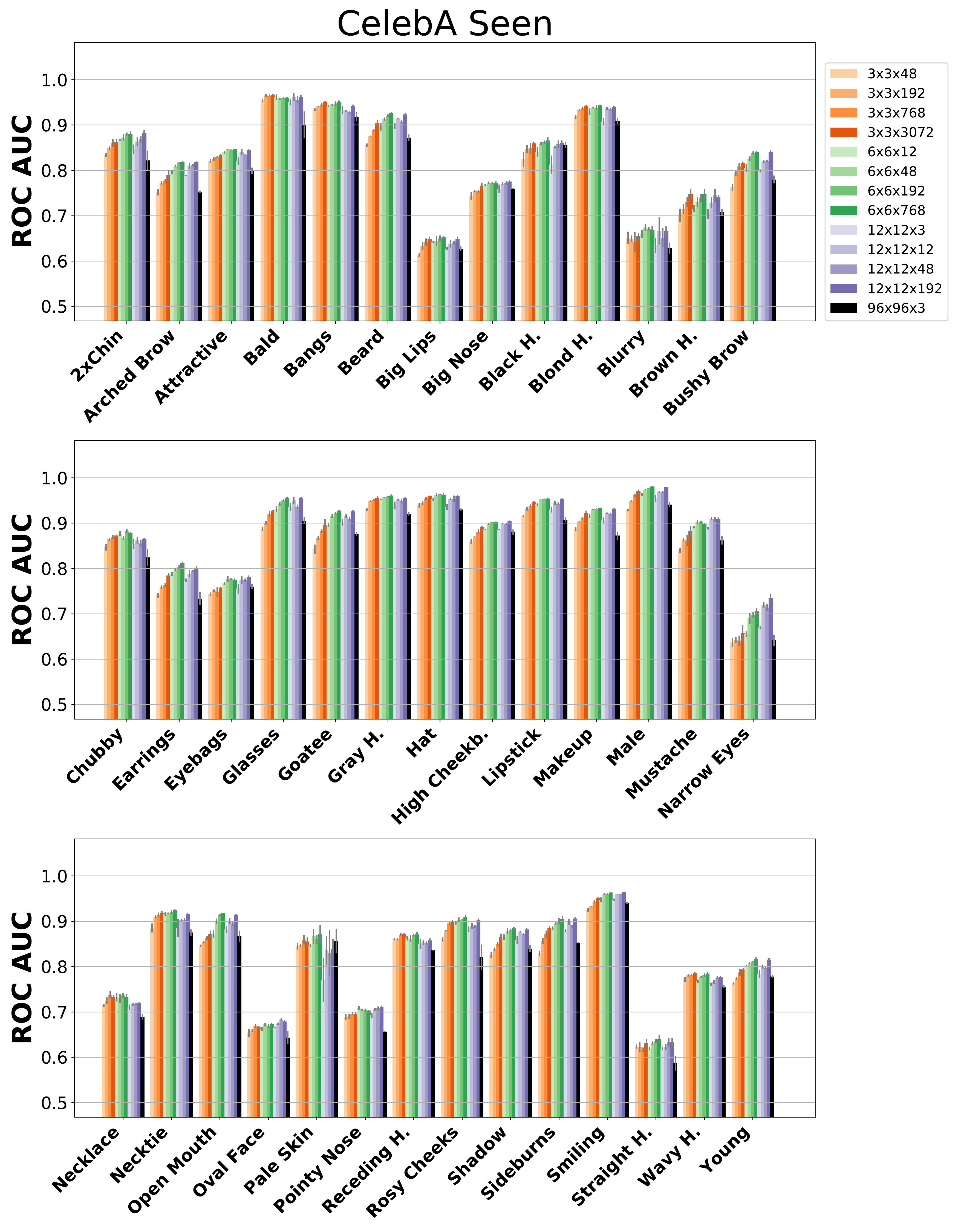}
	\caption{Classification performance by class for the CelebA dataset. The classifiers were trained on representations from samples that the CAE has seen during training. Configurations that have a common feature map size share the same color in the bar plot. Color intensity represents the amount of channels in the bottleneck (darker = more channels). Classification based on the raw inputs (baseline) is shown in black. Error bars indicate standard deviation over 9 runs (3 CAE seeds $\times$ 3 classifier seeds), except for the baseline where only the 3 classifier seeds are considered.}
\end{figure}

\begin{figure}[h]
	\centering
	\includegraphics[width=\subwidth]{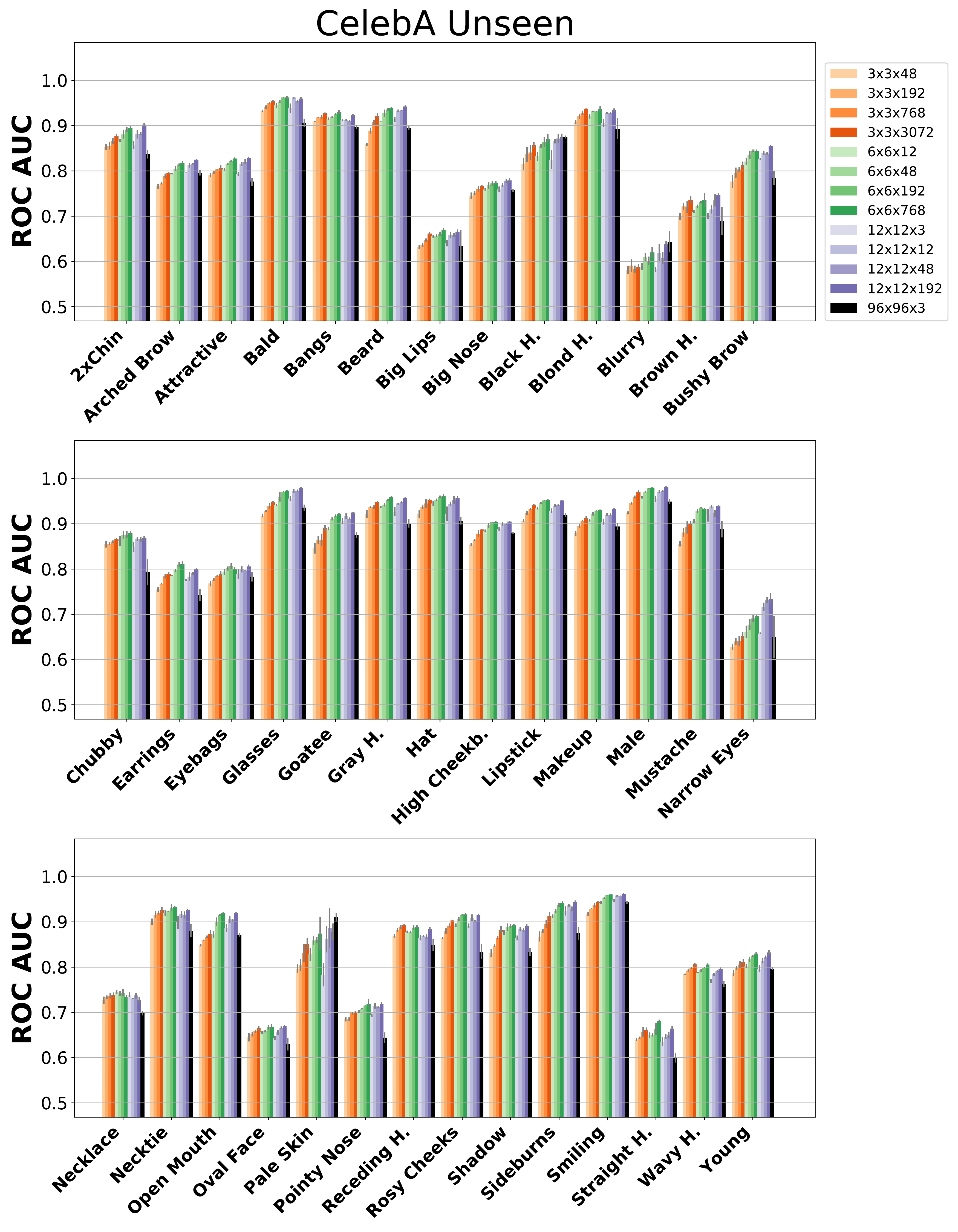}
	\caption{Classification performance by class for the CelebA dataset. The classifiers were trained on representations from samples that the CAE has not seen during training. Configurations that have a common feature map size share the same color in the bar plot. Color intensity represents the amount of channels in the bottleneck (darker = more channels). Classification based on the raw inputs (baseline) is shown in black. Error bars indicate standard deviation over 9 runs (3 CAE seeds $\times$ 3 classifier seeds), except for the baseline where only the 3 classifier seeds are considered.}
\end{figure}

\begin{figure}[h]
	\centering
	\includegraphics[width=\subwidth]{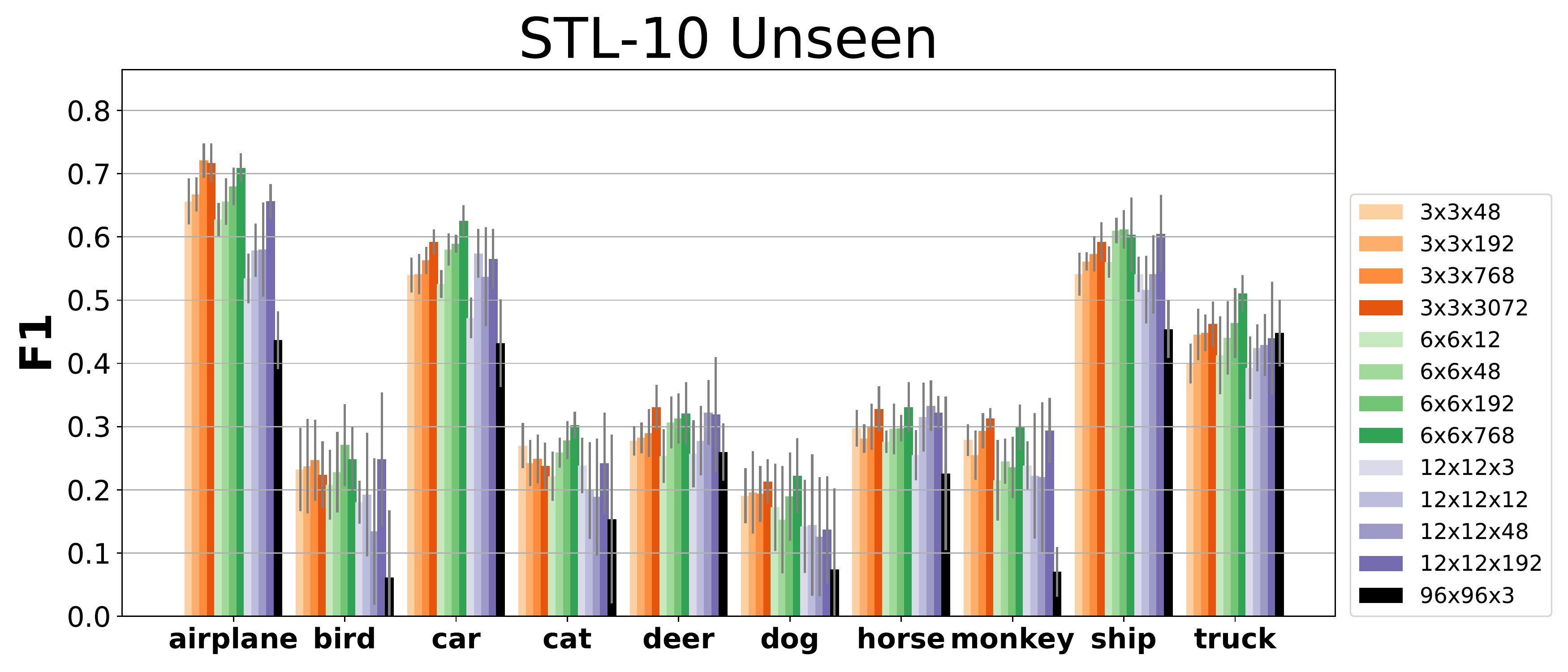}
	\caption{Classification performance by class for the STL-10 dataset. The classifiers were trained on representations from samples that the CAE has not seen during training. Configurations that have a common feature map size share the same color in the bar plot. Color intensity represents the amount of channels in the bottleneck (darker = more channels). Classification based on the raw inputs (baseline) is shown in black. Error bars indicate standard deviation over 9 runs (3 CAE seeds $\times$ 3 classifier seeds), except for the baseline where only the 3 classifier seeds are considered.}
\end{figure}

\renewcommand{\subwidth}{0.4\linewidth}
\begin{figure}[ht]
	\centering
	\subfloat{\includegraphics[width=\subwidth]{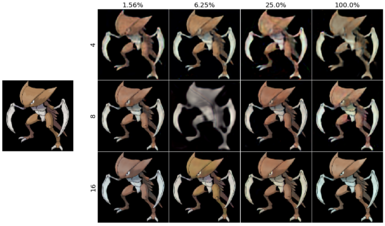}}\hspace{10pt}
	\subfloat{\includegraphics[width=\subwidth]{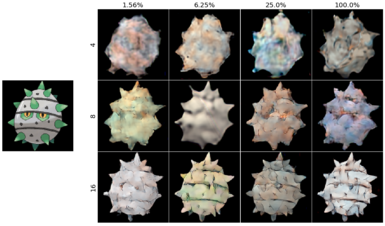}}\\
	\vspace{-2pt}
	\subfloat{\includegraphics[width=\subwidth]{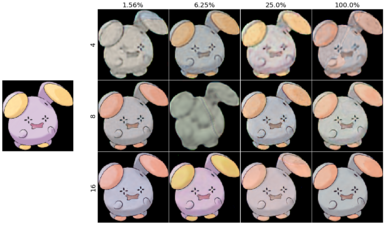}}\hspace{10pt}
	\subfloat{\includegraphics[width=\subwidth]{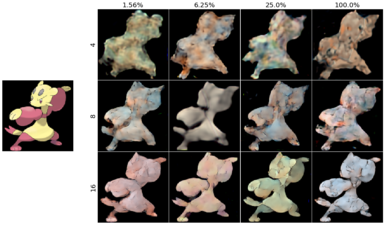}}\\
	\vspace{-2pt}
	\subfloat{\includegraphics[width=\subwidth]{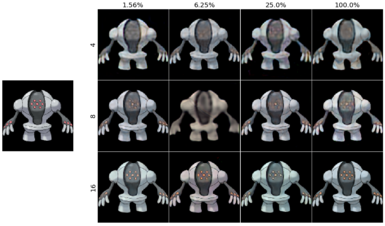}}\hspace{10pt}
	\subfloat{\includegraphics[width=\subwidth]{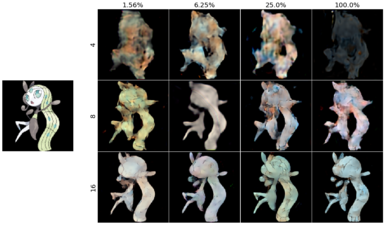}}\\
	\vspace{-2pt}
	\subfloat{\includegraphics[width=\subwidth]{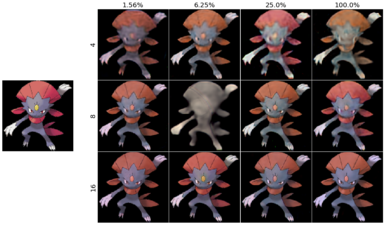}}\hspace{10pt}
	\subfloat{\includegraphics[width=\subwidth]{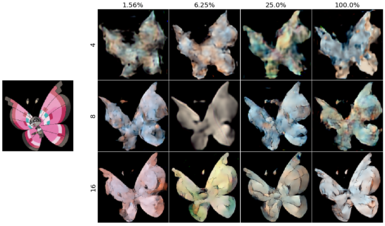}}\\
	\vspace{-2pt}
	\subfloat{\includegraphics[width=\subwidth]{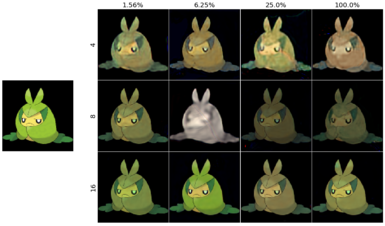}}\hspace{10pt}
	\subfloat{\includegraphics[width=\subwidth]{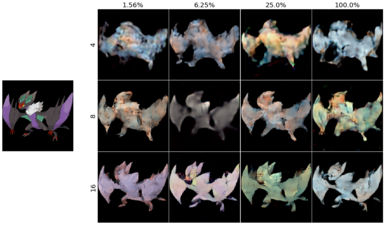}}\\
	\caption{Reconstructions of randomly picked samples from the Pokemon dataset. The left column contains samples from the training data, while on the right, we show samples from the test data. In each subfigure, the rows correspond to CAEs with the same bottleneck size (height, width), increasing from top to bottom. The columns group CAEs by the number of channels in the bottleneck, expressed as percentage relative  to input given bottleneck size. The image to the left of each grid is the input image.} \label{samples-pokemon}
\end{figure}

\begin{figure}[t]
	\centering
	\subfloat{\includegraphics[width=\subwidth]{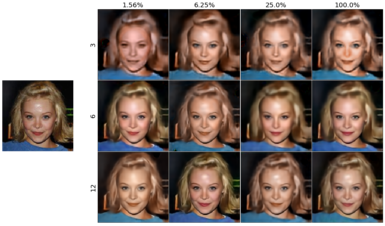}}\hspace{10pt}
	\subfloat{\includegraphics[width=\subwidth]{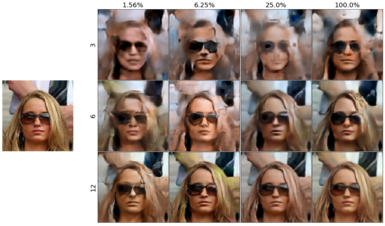}}\\
	\vspace{-2pt}
	\subfloat{\includegraphics[width=\subwidth]{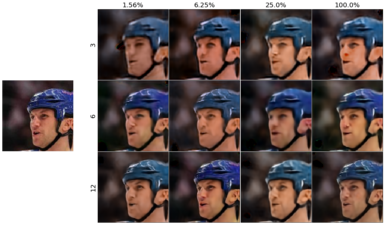}}\hspace{10pt}
	\subfloat{\includegraphics[width=\subwidth]{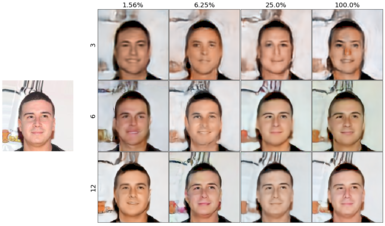}}\\
	\vspace{-2pt}
	\subfloat{\includegraphics[width=\subwidth]{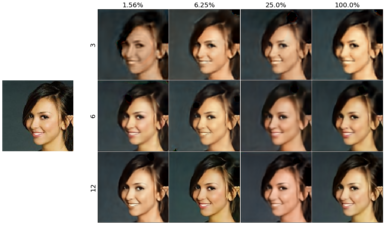}}\hspace{10pt}
	\subfloat{\includegraphics[width=\subwidth]{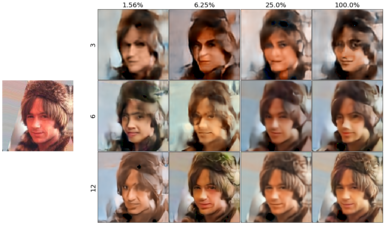}}\\
	\vspace{-2pt}
	\subfloat{\includegraphics[width=\subwidth]{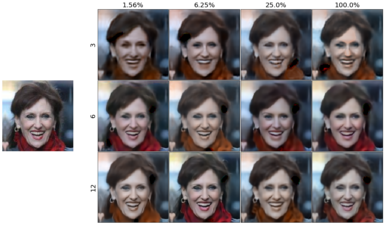}}\hspace{10pt}
	\subfloat{\includegraphics[width=\subwidth]{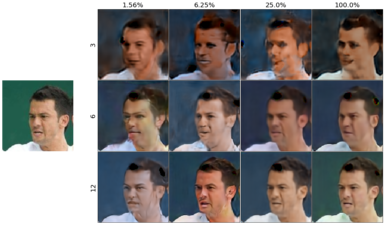}}\\
	\vspace{-2pt}
	\subfloat{\includegraphics[width=\subwidth]{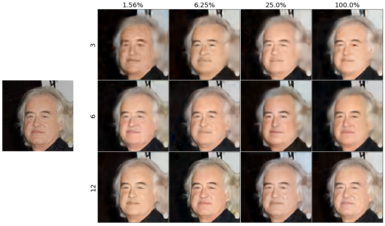}}\hspace{10pt}
	\subfloat{\includegraphics[width=\subwidth]{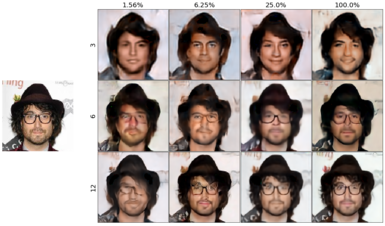}}\\
	\caption{Reconstructions of randomly picked samples from the CelebA dataset. The left column contains samples from the training data, while on the right, we show samples from the test data. In each subfigure, the rows correspond to CAEs with the same bottleneck size (height, width), increasing from top to bottom. The columns group CAEs by the number of channels in the bottleneck, expressed as percentage relative  to input given bottleneck size. The image to the left of each grid is the input image.} \label{samples-celeba}
\end{figure}

\begin{figure}
	\centering
	\subfloat{\includegraphics[width=\subwidth]{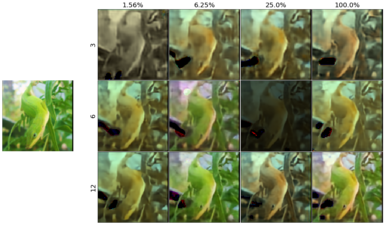}}\hspace{10pt}
	\subfloat{\includegraphics[width=\subwidth]{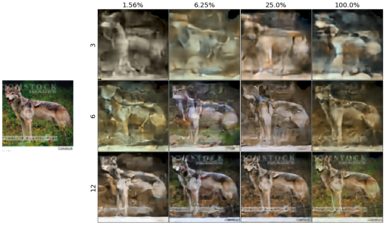}}\\
	\subfloat{\includegraphics[width=\subwidth]{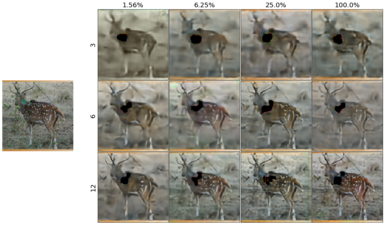}}\hspace{10pt}
	\subfloat{\includegraphics[width=\subwidth]{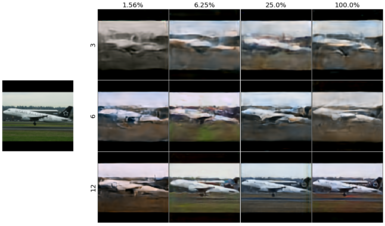}}\\
	\subfloat{\includegraphics[width=\subwidth]{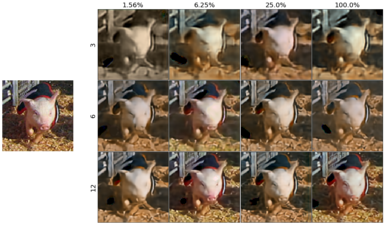}}\hspace{10pt}
	\subfloat{\includegraphics[width=\subwidth]{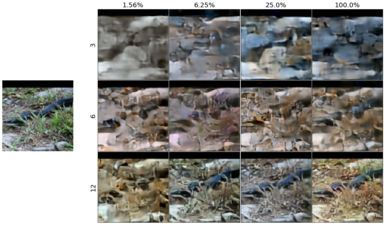}}\\
	\subfloat{\includegraphics[width=\subwidth]{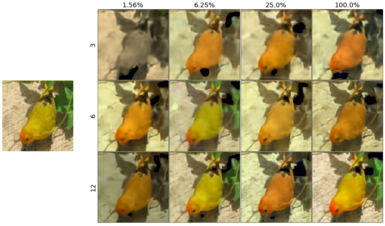}}\hspace{10pt}
	\subfloat{\includegraphics[width=\subwidth]{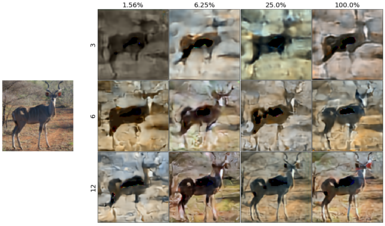}}\\
	\subfloat{\includegraphics[width=\subwidth]{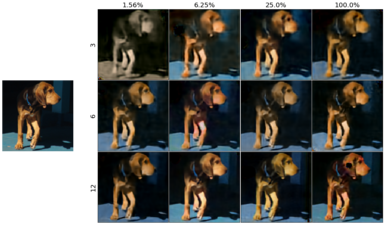}}\hspace{10pt}
	\subfloat{\includegraphics[width=\subwidth]{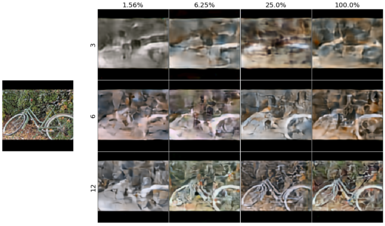}}\\
	\caption{Reconstructions of randomly picked samples from the STL-10 dataset. The left column contains samples from the training data, while on the right, we show samples from the test data. In each subfigure, the rows correspond to CAEs with the same bottleneck size (height, width), increasing from top to bottom. The columns group CAEs by the number of channels in the bottleneck, expressed as percentage relative  to input given bottleneck size. The image to the left of each grid is the input image.} \label{samples-stl10}
\end{figure}

\end{document}